\PassOptionsToPackage{table,xcdraw}{xcolor}
\documentclass[lettersize,journal]{IEEEtran}
\usepackage{amsmath,amsfonts}
\usepackage{array}
\usepackage[caption=false,font=normalsize,labelfont=sf,textfont=sf]{subfig}
\usepackage{textcomp}
\usepackage{stfloats}
\usepackage{url}
\usepackage{verbatim}
\usepackage{graphicx}
\usepackage{cite}
\usepackage{booktabs}    
\usepackage{orcidlink}
\usepackage[ruled,vlined]{algorithm2e}
\usepackage{amsmath}

\usepackage{multirow}
\usepackage[table,xcdraw]{xcolor}
\definecolor{Gray}{gray}{0.9}

\makeatletter
\let\NAT@parse\undefined
\makeatother
\usepackage{hyperref}

\begin{document}
\title{Unveiling the Underwater World: CLIP Perception Model-Guided Underwater Image Enhancement}

\author{Jiangzhong Cao~\orcidlink{0000-0002-5156-3538},
        Zekai Zeng~\orcidlink{0000-0001-7706-8730},
        Xu Zhang~\orcidlink{0000-0001-7685-7500},
        Huan Zhang\textsuperscript{*}~\orcidlink{0000-0002-5507-4985},~\IEEEmembership{Member,~IEEE,}
        Chunling Fan~\orcidlink{0009-0003-6698-0072},
        Gangyi Jiang~\orcidlink{0000-0002-9830-7048},~\IEEEmembership{Senior member,~IEEE,}
        Weisi Lin~\orcidlink{0000-0001-9866-1947}, ~\IEEEmembership{Fellow,~IEEE}
\thanks{This work was in part supported by. \emph{(Corresponding author: Huan Zhang.)}}%

\thanks{Jiangzhong Cao, Zekai Zeng, Xu Zhang, and Huan Zhang are with the School of Information Engineering, Guangdong University of Technology, Guangzhou 510006, China (e-mail: cjz510@gdut.edu.cn; 2112203035@mail2.gdut.edu.cn; 2122103221@mail2.gdut.edu.cn; huanzhang2021@gdut.edu.cn).}%

\thanks{Chunling Fan is with the School of Electronic and Communication Engineering, Shenzhen Polytechnic University, Shenzhen 518107, China  (e-mail:   fanchunling@szpu.edu.cn)} 

\thanks{Gangyi Jiang is with the Faculty of
Information Science and Engineering, Ningbo University, Ningbo 315211, China (e-mail: jianggangyi@nbu.edu.cn)}

\thanks{Weisi Lin is with the School of Computer Science and Engineering, Nanyang
	Technological University, Singapore 639798 (e-mail: wslin@ntu.edu.sg).}}
\markboth{Journal of \LaTeX\ Class Files,~Vol.~14, No.~8, August~2021}%
{Shell \MakeLowercase{\textit{et al.}}: A Sample Article Using IEEEtran.cls for IEEE Journals}


\maketitle
\begin{abstract}
High-quality underwater images are essential for both machine vision tasks and viewers with their aesthetic appeal. However, the quality of underwater images is severely affected by light absorption and scattering. Deep learning-based methods for Underwater Image Enhancement (UIE) have achieved good performance. 
However, these methods often overlook considering human perception and lack sufficient constraints within the solution space. Consequently, the enhanced images often suffer from diminished perceptual quality or poor content restoration. To address these issues, we propose a UIE method with a Contrastive Language-Image Pre-Training (CLIP) perception loss module and curriculum contrastive regularization.
Above all, to develop a perception model for underwater images that more aligns with human visual perception, the visual semantic feature extraction capability of the CLIP model is leveraged to learn an appropriate prompt pair to map and evaluate the quality of underwater images. 
This CLIP perception model is then incorporated as a perception loss module into the enhancement network to improve the perceptual quality of enhanced images. Furthermore, the CLIP perception model is integrated with the curriculum contrastive regularization to enhance the constraints imposed on the enhanced images within the CLIP perceptual space, mitigating the risk of both under-enhancement and over-enhancement.
Specifically, the CLIP perception model is employed to assess and categorize the learning difficulty level of negatives in the regularization process, ensuring comprehensive and nuanced utilization of distorted images and negatives with varied quality levels.
Extensive experiments demonstrate that our method outperforms state-of-the-art methods in terms of visual quality and generalization ability.

\end{abstract}


\section{Introduction}
\IEEEPARstart{D}{uring} recent years, there has been a substantial growth of underwater images across various fields such as marine biology, underwater exploration, and underwater archaeology~\cite{paull2013auv,cong2021rrnet}. In these scenarios, the captured underwater images play a pivotal role not only in machine vision tasks like water body classification and ocean condition monitoring~\cite{li2019nested,wang2023reinforcement} but also in facilitating captivating visual experiences of the underwater world, such as heritage and landscapes~\cite{anwar2020diving}.
However, these underwater images typically exhibit a range of distortions, including color distortion, low contrast, and blurred details, primarily caused by the varying attenuation of light at different wavelengths and the scattering effects induced by marine microorganisms~\cite{schettini2010underwater}. Therefore, to obtain clearer and higher quality underwater images for machine vision tasks as well as human visual perception, effective Underwater Image Enhancement (UIE) is highly desired.

Existing methods for UIE can primarily be categorized into three major types: visual prior-based methods, physical model-based methods, and deep learning-based methods. Visual prior-based methods mainly adjust the pixel values of an image from the perspectives of contrast, brightness, and saturation to improve the image quality~\cite{ancuti2012enhancing},~\cite{iqbal2010enhancing}. But they do not consider the physical degradation process of underwater images. Physical model-based methods~\cite{iqbal2010enhancing,he2010single,li2016underwater} mainly estimate the physical medium parameters of the imaging process of underwater images. Due to the complexity of the underwater environment, it is difficult to adapt the assumed models to different types of underwater environments. Additionally, accurately estimating a large number of parameters poses a significant challenge, which results in poor robustness of such models. 

Recently, deep learning-based methods have shown excellent performance in many computer vision fields~\cite{zou2023object}, and researchers have started to design UIE networks in an end-to-end manner~\cite{li2019underwater,li2021underwater,islam2020fast,peng2023u,guo2023underwater}. Even though these methods have somewhat enhanced the quality of images, they still have their limitations. For example, Peng et al.~\cite{peng2023u} proposed a Transformer-based Ushape model to improve image quality, but the model has a large number of parameters and overlooks human visual perception, resulting in suboptimal visual perception effects in the enhanced images. Li et al.~\cite{guo2023underwater} proposed a quality assessment model, URanker, and incorporated it into the enhancement network NU2Net as a supervised loss to improve model performance. However, the enhanced image still exhibits some defects, e.g., reddish colors in the enhanced images, for the quality assessment model URanker may not well reflect the perceptual quality of underwater images. 
In addition, due to the unavailability of authentic ground-truth for underwater images, the ground-truth underwater images is subjectively selected by volunteers. They select the highest-scored image through multiple ratings and voting to serve as pseudo ground-truth, leading to a significantly human-subjective influence on the quality of pseudo ground-truth~\cite{li2019underwater}~\cite{peng2023u}. The mapping between distorted images and pseudo ground-truth in learning may introduce biases, potentially resulting in under-enhancement or over-enhancement of the processed underwater images.
Studies have shown that the rich visual language encapsulated in the Contrastive Language-Image Pre-Training (CLIP) model can be used to assess the quality and perception of images\cite{wang2023exploring}. Inspired by this, to better reflect and improve the perceptual quality of enhanced underwater images, a CLIP perception model is proposed that is more in line with human perception and can be applied to enhancement networks. Specifically, following the recent method~\cite{wang2023exploring}, a suitable initial antonymic prompt pair is selected to evaluate underwater image quality and fine-tune them accordingly. Through this fine-tuning process, the model's evaluation of underwater image quality becomes more consistent with human perception. 
Subsequently, it is incorporated into the enhancement network as a perception loss during model training, thereby ensuring that the enhanced images closely align with human perception.

Furthermore, in order to avoid under-enhancement or over-enhancement in enhancing the underwater images, we introduce more extra constraints on the solution space by proposing a curriculum contrastive regularization on the enhanced images using versatile negative examples. Compared to traditional contrastive learning~\cite{wu2021contrastive} that relies on negatives generated through random sampling, it can provide more effective constraints in the solution space. Specifically, the distorted input images and the images repaired using existing UIE methods are used as negatives. Then the CLIP perception model is employed as a scoring criterion, whereby negatives are categorized according to the degree of their distortion, with their weights being dynamically adjusted.


In summary, our contribution is as follows.
\begin{itemize}
\item We propose an underwater image enhancement method that incorporates vision model perception loss and curriculum contrastive regularization for training. These dual constraints can improve the performance of a UIE network in a plug-and-play manner without changing its network structure.
\item We introduce a CLIP perception model based on learnable antonymic prompt pairs that could more accurately reflect the perceptual quality of underwater images. This model is integrated into an enhancement network as a CLIP perception loss, guiding the network to enhance underwater images with improved perceptual quality.
\item We propose a curriculum contrastive regularization 
combined with CLIP, wherein the CLIP perception model serves as a criterion to dynamically adjust the difficulty levels of various negatives. Hence, the quality of underwater images could be enhanced perceptually under more precise constraints by exploiting the mutual relationship between versatile negatives and the anchor.
\end{itemize}

\section{Related Work}
\subsection{Deep Learning-based Underwater Image Enhancement}
UIE has attracted considerable attention and research efforts as a crucial step in improving the visual quality of underwater images. Recently, deep learning-based methods have gained significant attention in UIE research. 
These methods mainly fall into categories such as Convolutional Neural Networks (CNN)-based~\cite{li2019underwater,li2021underwater,guo2023underwater}, Generative Adversarial Networks (GAN)-based~\cite{islam2020fast},~\cite{10155564} and Transformer-based~\cite{9881581},~\cite{peng2023u} frameworks. 
Li et al.~\cite{li2019underwater} proposed a CNN-based model  Waternet, in which various modes of enhanced inputs were integrated with their corresponding confidence maps to fuse final enhanced images. 
Li et al.~\cite{li2021underwater} introduced Ucolor, a multi-color spatial embedding UIE network, utilizing medium transmission guidance to integrate features from various color spaces into the network architecture. 
Liu et al.~\cite{10192442} proposed a multiscale dual-color space UIE network called MSDC-Net. Some researchers began to apply GAN to UIE tasks for its powerful data distribution modeling and image generation ability. Islam et al.~\cite{islam2020fast} proposed a real-time UIE network called FUnIE-GAN based on conditional GAN. Yan et al.~\cite{10251985} leveraged the strength of CycleGAN and incorporated a physical model-driven strategy by predicting the inherent information of underwater images, e.g., transmittance maps, ensuring visually pleasing and physically realistic underwater images.
Cong et al.~\cite{10155564} proposed a UIE method called PUGAN, which initially trains a physical model and then incorporates the generated color-enhanced underwater images as auxiliary information into subsequent GAN networks. 
Recently, Transformer-based UIE methods have emerged and achieved good performance for Transformer's global modeling ability.
Huang et al.~\cite{9825662} proposed a Swin Transformer network based on multiscale cascade modules and channel attention mechanism, and proposed an adaptive group attention mechanism to reduce the attention learning parameters of the UIE network. Peng et al.~\cite{peng2023u} specifically designed both channel and spatial attention modules in the transformer-based UIE model to address the severe local distortion and color artifacts in the underwater images.
Lately, some methods have incorporated quality assessment models directly into UIE networks, either as a network component or part of the loss function to guide the enhancement process.
Li et al.~\cite{li2022human} proposed a UIE network utilizing the Underwater Image Quality Assessment Network (UIQAN) to boost the performance in both two ways. Li et al.~\cite{guo2023underwater} integrated the underwater image quality assessment method Uranker into the UIE network as a part of the loss function. 

Generally, the aforementioned UIE methods have achieved fairly good performance. However, the respective enhanced underwater images are still not perceptually pleasant. Although some approaches have integrated quality assessment models into UIE networks, further research on perception guidance is needed. In addition, the pseudo-ground truth underwater images may bias the enhanced underwater images over- or under-enhancement. Therefore, more constraints should be imposed on the solution space of the enhanced underwater images beyond solely relying on the pseudo-ground truth. 



\subsection{CLIP and Prompting in Vision}
CLIP has achieved promising results in zero-shot classification using large-scale image-text pair datasets~\cite{radford2021learning}. It has demonstrated its generality in many high-level researchs~\cite{zang2022open,lin2024mirrordiffusion,kuo2022f,zhou2022extract}. And prompt learning as a core component of vision-language models, has received significant attention and research recently. For example, research works such as CoOp~\cite{zhou2022learning}, CoCoOp~\cite{zhou2022conditional}, integrate prompt learning into the vision-language model CLIP to enhance its adaptability in downstream visual tasks. Recent studies~\cite{wang2023exploring,zhang2023blind,10378578} have shown that the rich visual semantic information in CLIP can be also utilized for low-level vision tasks. For example, Wang et al.~\cite{wang2023exploring} leveraged the visual semantic feature extraction capability of the CLIP model and proposed a quality assessment model using an antonymous prompt pair to evaluate the perceptual and abstract quality of an image. Li et al.~\cite{10378578} integrate CLIP into the backlit image enhancement network and used an iterative prompt learning strategy to generate more precise prompts, further improving model performance. Additionally, many underwater image quality assessment models~\cite{yang2021reference},~\cite{jiang2022underwater} struggle to accurately represent image distortion degrees and cannot be integrated into enhancement networks due to their non-differentiability. In this paper, we propose a CLIP perception model for the UIE task, which can assess underwater images more closely aligned with human perception and can be integrated into enhancement networks to improve the perceptual quality of the results.


\subsection{Contrastive Learning}
Contrastive learning has been widely used for high-level visual tasks~\cite{chen2020simple,grill2020bootstrap,lin2022exploring,he2020momentum}. The main idea is to continuously pull an anchor point closer to a positive point while pushing away from a negative point through contrastive loss. Recently, this approach has also been applied to low-level visual problems~\cite{zheng2023curricular,wu2021contrastive,huang2023contrastive}. Han et al.~\cite{han2022underwater} proposed an unsupervised UIE method that employs contrastive learning. However, due to the use of a non-reference dataset, the content of negatives may differ from those of positives, and their embeddings may be too far apart in the latent feature space, leading to ineffective constraints on the underwater images. Huang et al.~\cite{huang2023contrastive} proposed a semi-supervised UIE method utilizing contrastive learning, in which well-enhanced underwater images evaluated by general non-reference metrics were selected and utilized as pseudo-positives. Due to potential inaccuracies in evaluation, certain results with inadequate perceptual quality may be misclassified as pseudo-positives, possibly disrupting the process of model training. 
In the field of dehazing, Wu et al.~\cite{wu2021contrastive} proposed contrastive regularization (CR) to train models using feature information from negatives and positive. Then, Zheng et al.~\cite{zheng2023curricular} introduced the concept of consensus negatives to provide more effective constraints in the solution space based on CR and proposed curriculum contrastive regularization. Inspired by ~\cite{zheng2023curricular}, we propose a curriculum contrastive regularization method that combines a perception model for evaluating perceptual scores of enhanced image and negatives to train the enhancement networks, where reference images are used as positives and images with different distortion levels with the same content are used as negatives. This scheme could facilitate the imposition of more effective constraints on the UIE model's solution space, thus avoiding under- or over-enhancement of the enhanced underwater images.

\section{Method}
\subsection{Overview}
\begin{figure*}
\centering
\includegraphics[width=1\linewidth]{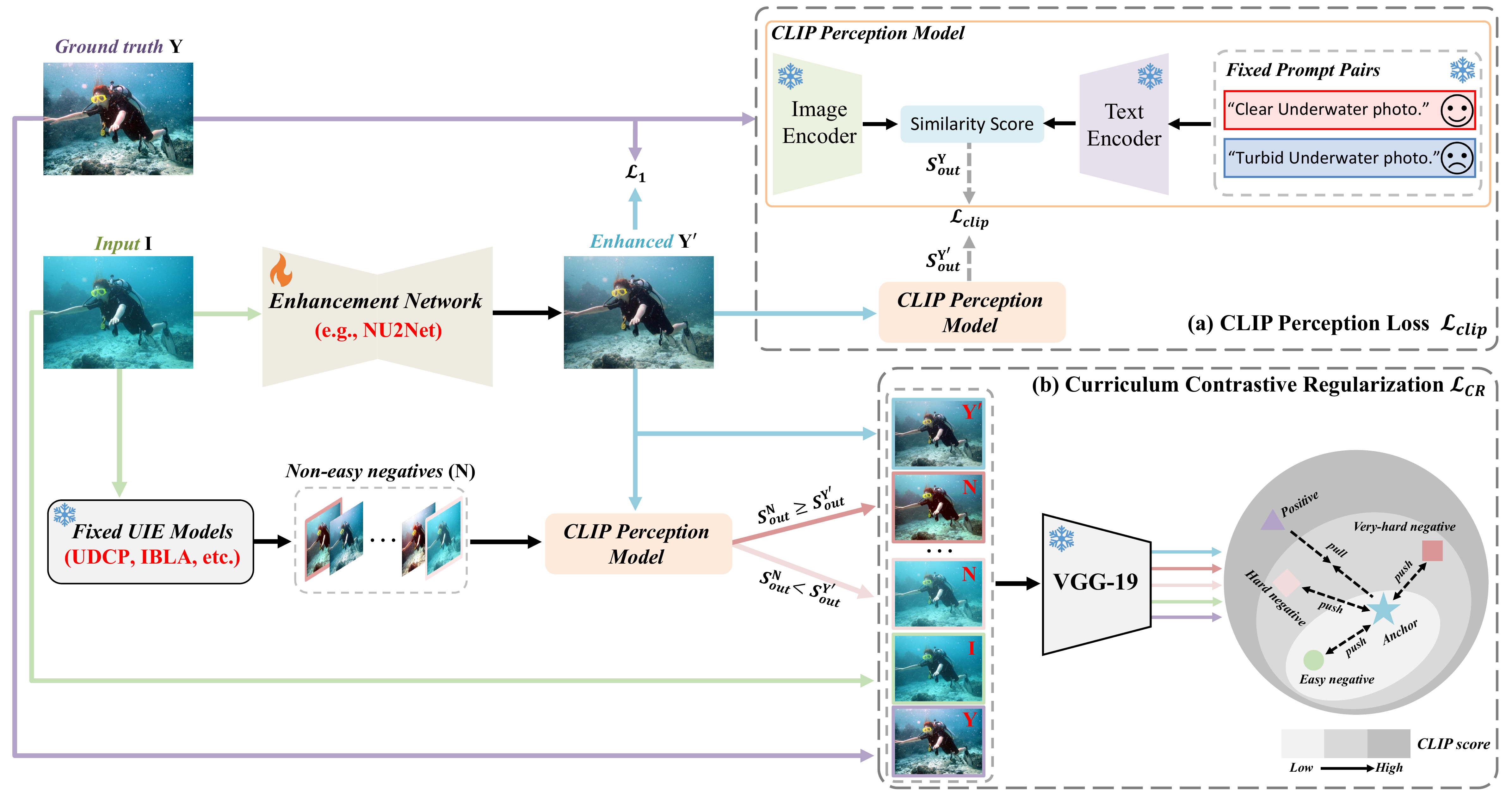}
\caption{\label{fig:UIE}The proposed framework of UIE network. (a) The CLIP Perception Loss module $\mathcal{L}_{clip}$. (b) The curriculum contrastive regularization module $\mathcal{L}_{CR}$. Our enhancement network utilizes modules $\mathcal{L}_{clip}$, $\mathcal{L}_{CR}$ and $\mathcal{L}_1$ as supervised loss conditions, ultimately leading to visually desirable results.}
\end{figure*}

The framework of our proposed method is demonstrated in Fig. \ref{fig:UIE}, which involves two stages, i.e., the CLIP perception model and the overall enhancement network. The procedure is to first learn a CLIP perception model and then the learned CLIP perception model is integrated into the enhancement network as the CLIP perception loss module (a) and the curriculum contrastive regularization module (b). Further details of the framework of the CLIP perception model can be seen in Fig. \ref{fig:CLIP}.


\subsection{CLIP Perception Model}

\begin{figure}
\centering
\includegraphics[width=1\linewidth]{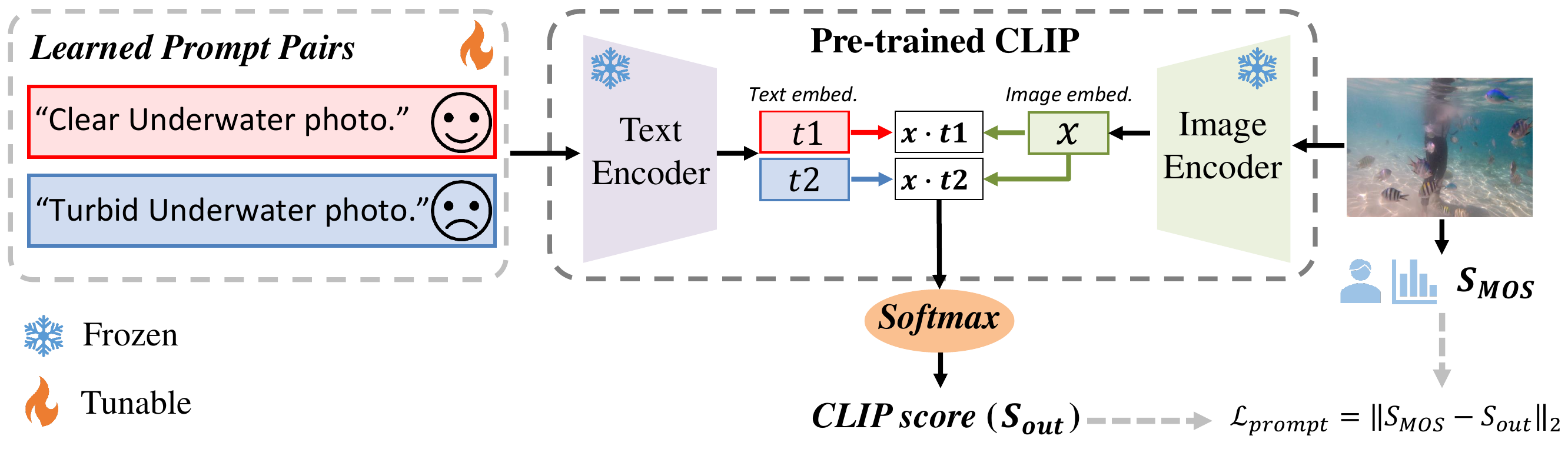}
\caption{\label{fig:CLIP}The framework of CLIP perception model.}
\end{figure}

In previous CLIP-based works ~\cite{zhou2022extract}~\cite{rao2022denseclip}, typically only a single text prompt was utilized to train a CLIP model.
However, this approach is not suitable in the field of image quality perception due to the semantic ambiguity present in the text~\cite{khurana2023natural}. The same text may have different meanings without contexts (e.g., ``clean house image'' can represent either images of a clean house or images related to the action of cleaning the house). Similar to the work in ~\cite{wang2023exploring}, we use a learnable antonymic prompts pair to address this issue. As shown in Fig. \ref{fig:CLIP}, a prompt that describes a high-quality underwater image is defined as a positive prompt, and conversely, a prompt describing a distorted underwater image is defined as a negative prompt. Since the choice of initial prompts would influence the effectiveness of the model, as indicated in~\cite{wang2023exploring}, the positive and negative prompts are discreetly initialized as [``Clear Underwater photo.'', ``Turbid Underwater photo.'']. 

Based on the initial prompts, the prompt in the CLIP perception model will be learned and updated. First, the underwater image is fed into the image encoder $\Phi_{image}$ of the CLIP model to get its latent encoding while the positive and negative prompts are input into the text encoder $\Phi_{text}$ to generate their respective latent encodings. Then, two scores, i.e., the score ${S_p}$ representing the similarity between the underwater image and positive prompt (Eq.\ref{eq1}), and the score ${S_n}$  representing the similarity between the underwater image and negative prompt (Eq.\ref{eq2}), are obtained by calculating the text-image similarity in CLIP latent space. And our model finally outputs the processed ${S_p}$ as the predicted quality score ${S_{out}}$ for the input image by applying the softmax function to the vector (${S_p}$, ${S_n}$) (Eq.\ref{eq3}). Finally, the Mean Squared Error (MSE) between the predicted quality score  ${S_{out}}$ and the corresponding image label ${S_{MOS}}$ in Eq.(\ref{eq4}) is employed as the loss function for learning the optimal prompts. The process of the prompt learning is described as
\begin{algorithm}
\SetAlgoLined
\DontPrintSemicolon
\SetNlSkip{1em}
\KwIn {Image $\mathbf{I}$ with MOS scores, Initial Prompts Pair $\mathbf{T}^0_{\phi}$, $\phi \in \{{p,n}\}$, the number of epochs M}
\KwOut{Image CLIP perception score $S_{out}$, the final learned prompts $\mathbf{T}_{\phi}$}
\For{\(j = 1\) \KwTo \(M\)}
    {
     $\mathbf{T}^{j+1}_{\phi} \leftarrow \mathbf{T}^{j}_{\phi}$\;
     $\Phi_{image} = \text{CLIP Image Encoder}(\mathbf{I})$\;
     $\Phi_{text} = \text{CLIP Text Encoder}(\mathbf{T}^j_{\phi})$\;
     Calculate cosine similarity between $\Phi_{image}$ and $\Phi_{text}$ (Eq.\ref{eq1},\ref{eq2})\;
     After applying softmax, output the score that represents image quality $S_{out}$ (Eq.\ref{eq3})\;
     Compute $\mathcal{L}_2$ between $S_{out}$ and $S_{MOS}$ (Eq.\ref{eq4})\;
     Backpropagation updates the Prompts\;  
     }
 return $\mathbf{T}_{\phi} = \mathbf{T}^{M}_{\phi}$
\caption{Training Procedure for CLIP Perception Model}
\label{alg:algorithm_1}
\vspace{-0em}
\end{algorithm}
\setlength{\parskip}{0.2cm plus4mm minus3mm}
\begin{align}
    \label{eq1} &{S_{p}} =\cos(\Phi_{image}(\mathbf{I}),\Phi_{text}(\mathbf{T}_{p})),  \\
    \label{eq2} &{S_{n}} =\cos(\Phi_{image}(\mathbf{I}),\Phi_{text}(\mathbf{T}_{n})) , \\
    \label{eq3} &{S_{out} =\frac{e^{S_{p}} }{e^{S_{p}}+e^{S_{n}}}} , \\
    \label{eq4} &\mathcal{L}_{prompt} = {||S_{MOS}-S_{out}||_2},  
\end{align}
where $ \mathbf{T}_{\phi} \in \mathbb{R}^{N \times 512}$, $\phi \in \{p, n\},  $  $\mathbf{T}_p$ represents positive prompt, and $\mathbf{T}_n$ represents negative prompt, $N$ represents the number of embedded tokens in each prompt, $\mathbf I \in \mathbb{R}^{H\times W\times 3}$ represents the input image, $\left \| \cdot  \right \| _2$  represent the $\mathcal{L}_2$ loss, and ${S_{MOS}}$ represents the corresponding Mean Opinion Score (MOS) of the image, and larger ${S_{MOS}}$ denotes higher image quality. 
From Eqs.(\ref{eq1})-(\ref{eq4}), it can be inferred that images with higher $S_{MOS}$ tend to have higher $S_{out}$. The detailed training process for the CLIP model is provided in Algo.\ref{alg:algorithm_1}. Through prompt learning, the CLIP perception model has been trained to achieve the highest Pearson's Linear Correlation Coefficient (PLCC) and the second-highest Spearman's Rank-Order Correlation Coefficient (SROCC) on the test set of the UEQAB dataset~\cite{li2022human}, as depicted in Table \ref{tab:table1}. This indicates that our proposed CLIP perception model outputs scores that align more closely with the subjective scores, indicating that our CLIP perception model can reflect the perceptual quality of underwater more accurately.

\begin{algorithm}
\SetAlgoLined
\DontPrintSemicolon
\SetNlSkip{1em}
\KwIn{Distorted underwater image \(\mathbf{I}\), Ground truth \(\mathbf{Y}\)}
\KwOut{Enhanced image \(\mathbf{Y^{'}}\), learned UIE network $F$ with parameters $\Theta$ \ }
\textbf{Initialization}: fixed CLIP perception model $Q$,  number of non-easy negatives \(z\), fixed UIE models $\Psi$ for generating non-easy negatives, \(epochs = 800\) \;

\tcp{Phase 1: Obtain non-easy negatives}
\For{\(q = 1\) \KwTo \(z\)}{
    \(\mathbf{N}_q = \Psi_q(\mathbf{I})\)
}

\tcp{Phase 2: Train Enhancement Network}
\For{\(j = 1\) \KwTo \(epochs\)}{
    \(\mathbf{Y^{'}} = F(\mathbf{I};\Theta_j)\) \;
    \(S_{out}^{\mathbf{Y^{'}}} = Q(\mathbf{Y^{'}})\), \(S_{out}^{\mathbf{Y}} = Q(\mathbf{Y})\), \(S_{out}^{\mathbf{N}_{q}} = Q(\mathbf{N_{q}})\) \;
    
    \tcp{Divide non-easy negatives by $S_{out}^{\mathbf{N}_{q}}$}
    \uIf{\(S_{out}^{\mathbf{Y^{'}}} > S_{out}^{\mathbf{N}_{q}}\)}{
        Assign \(\mathbf{N}_{q}\) as Hard negatives \;
    }
    \Else{
        Assign \(\mathbf{N}_{q}\) as Very-hard negatives \;
    }
    Assign different weights to \(\mathbf{N}_{q}\) and compute curriculum contrastive regularization \(\mathcal{L}_{CR}\) (Eq.\ref{eq6}, \ref{eq7}) \;
    Compute \(\mathcal{L}_1\) loss \;
    Compute CLIP perception loss \(\mathcal{L}_{clip}\) (Eq. \ref{eq5}) \;
    Obtain total loss \(\mathcal{L}_{total}\) (Eq. \ref{eq8}) \;
    Backpropagation to update network parameters \;
}
\caption{Training Procedure for Enhancement Network}
\label{alg:algorithm_2}
\end{algorithm}
\vspace{-0.5em}

\subsection{Perception Loss with CLIP Model}
Existing deep learning-based methods for UIE primarily focus on minimizing the $\mathcal{L}_1$ or $\mathcal{L}_2$ loss between the enhanced result and the reference image. However, these methods fail to consider human visual perception, resulting in poor perceptual quality of the result.
To address this issue, several approaches have proposed incorporating additional perception loss into the loss function. For instance, Li et al.~\cite{guo2023underwater} first train an underwater image quality perception model called Ranker, and then transform this model as Ranker loss to train a UIE model. Additionally, Li et al.~\cite{li2022human} integrate their proposed quality assessment module, UIQAN, into a UIE model as a perception loss, which improves the perceptual quality of the enhanced underwater images. Motivated by these methods, we also propose a perception loss function leveraging the CLIP perception model in a UIE model. As shown in Fig. \ref{fig:UIE}, when integrating the CLIP perception model into the loss module (a), the weights and learned prompts of the CLIP model remain fixed. Within the UIE network, the input image $\mathbf{I}$ is fed into the enhancement network and obtained the enhanced image $\mathbf{Y^{'}}$. Subsequently, the enhanced image $\mathbf{Y^{'}}$ and the Ground Truth $\mathbf{Y}$ are both processed through the CLIP perception model to yield the scores $S_{out}^{\mathbf{Y^{'}}}$ and $S_{out}^{\mathbf{Y}}$, respectively. Similar to ~\cite{li2022human}, the loss function incorporating CLIP perception model is formulated as
\begin{align}
\label{eq5}
\mathcal{L}_{clip}=max(0,((1 - S_{out}^{\mathbf{Y^{'}}})-\alpha (1-S_{out}^{\mathbf{Y}}))),
\end{align}
where $\alpha\in [0,1]$ denotes the hyperparameter used to regulate the desired level of quality for the enhanced images generated by the UIE model. When $\alpha < 1.0$, it encourages the enhanced images to exhibit perceptual quality surpassing that of the reference images. For our experiments, $\alpha$ is set as 0.975.

\subsection{Curriculum Contrastive Regularization with CLIP Model}

In conventional contrastive regularization for low-level image vision tasks, the image content of multiple negatives often does not align with the anchor (the image to be solved or enhanced), resulting in a significant divergence in their mapping distance within the feature space. In this case, the contrast between the anchor and negatives fails to provide satisfactory constraints on the solution space, thus leading to the erratic quality of the final enhanced image. To avoid this problem, a curriculum contrastive regularization for image dehazing was proposed in~\cite{zheng2023curricular}, in which multiple non-easy negatives were obtained by various dehazing methods from the same input, thus sharing the same content with the anchor. In addition, the non-easy negatives of varying quality levels could expedite model convergence and reduce the potential learning uncertainty.

Inspired by~\cite{zheng2023curricular}, the curriculum contrastive regularization is leveraged in our method for the UIE task. As demonstrated in module (b) in Fig. \ref{fig:UIE}, the enhanced image $\mathbf{Y^{'}}$ from the network is regarded as the anchor, the reference image $\mathbf{Y}$ as the positive, the input image $\mathbf{I}$ as the easy negative, and the enhanced images $\mathbf{N}_{q}$ obtained from the existing UIE method (e.g., UDCP~\cite{drews2013transmission}, IBLA~\cite{peng2017underwater}) as non-easy negatives (hard negatives or very-hard negatives). The less severe degree of image distortion indicates the much more difficulty of negatives. However, the complexity and variability inherent in underwater image distortion render many conventional image quality assessment metrics (such as PSNR and UIQM) ineffective. Consequently, there arises a pressing need for a more precise and effective evaluation criterion to discern hard or very-hard negatives from non-easy negatives. 

Based on the preceding analysis, the CLIP perception model, which is capable of more accurately measuring the quality of underwater images, is leveraged to act as the division criterion for the degree of negatives in the curriculum contrastive regularization. Specifically, in the $i$-th epoch, if the CLIP score $S_{out}^{\mathbf{N}_{q}}$ of a non-easy negative $\mathbf{N}_{q}$  is greater than the CLIP score $S_{out}^{\mathbf{Y^{'}}}$ of the enhanced image $\mathbf{Y^{'}}$, it is classified as a very-hard negative; otherwise, it is classified as a hard negative. For different types of non-easy negatives, the weights for them in the $i$-th epoch are set as follows: 
\begin{equation}
\label{eq6}
W_{t}(\mathbf{N}_{q})=\left\{\begin{array}{l}
1 +\gamma, \quad S_{out}^{\mathbf{Y^{'}}}>S_{out}^{\mathbf{N}_q},  \\
1-\gamma, \quad \rm{otherwise}, 
\end{array}\right.
\end{equation}
where $\mathbf{N}_{q}$ represents non-easy negatives, $q = 1,2,\cdots,z$, $z$ is the number of the non-easy negatives and is set to 6. Since non-easy negatives are closer to positive samples, uncertainty may be introduced into the model's learning process occasionally. For the anchor $\mathbf{Y^{'}}$, to prioritize the positive force in regularization and mitigate potential learning uncertainty caused by non-easy negatives, the weights assigned to hard and very-hard negatives are set to $1+\gamma$ and $1-\gamma$, respectively. $\gamma$ is set to 0.25.
As the training progresses, the quality of the enhanced image $\mathbf{Y^{'}}$ improves, leading to higher CLIP perception scores. Consequently, these scores surpass those of the very-hard negatives, causing the very-hard negatives to transition into hard negatives. Thus, the allocation weights for these negatives undergo dynamic adjustments.
In addition, the weight assigned to the easy negative remains constant and is greater than that of non-easy negatives. Specifically, it is set to the quantity of non-easy negatives, namely $z$.

The curriculum contrastive regularization $\mathcal{L}_{CR}$ can thus be formulated as follows
\begin{equation}
\label{eq7}
\mathcal{L}_{CR} =\sum_{i=1}^{n} \xi _{i}\frac{\mathbf E_{i}(\mathbf{Y,Y^{'}})} { {\textstyle \sum_{q=1}^{z}} {W_{t}(\mathbf N_{q}) \mathbf{E}_{i}(\mathbf{N}_q, \mathbf{Y^{'}})} +z\cdot \mathbf{E}_{i}(\mathbf{I,Y^{'}})}
\end{equation}
where $\mathbf E_{i}(\mathbf{X,Y})=\left \|\mathbf{V}_{i}(\mathbf X) -\mathbf V_{i}(\mathbf Y)  \right \|_{1}$, $\mathbf{V}_{i}(\cdot)$, $i=1,2,\cdots,n$ represents the $i$-th hidden feature extracted from the pre-trained VGG-19, $||\cdot||_1$ represent the $\mathcal{L}_1$ distance, and $\left \{ \xi_i \right \}$ represents the corresponding set of weight hyperparameters. Compared to conventional contrastive regularization approaches, more advantages can be offered by (Eq.\ref{eq7}) in our proposed method. It leverages not only the features of the input distorted images but also incorporates the features of images enhanced through various UIE methods, thus imposing sufficient constraints on the underwater images to be enhanced. 

Finally, our UIE loss function is shown below:
\begin{equation}
\label{eq8}
\mathcal{L}_{total}=\mathcal{L}_1+\lambda _1 \mathcal{L}_{clip}+\lambda _2 \mathcal{L}_{CR} ,
\end{equation}
where $\lambda _1$ and $\lambda _2$ are parameters to make a balance among the three loss items. And the detailed training process of the UIE network is provided in Algo.\ref{alg:algorithm_2}.


\section{Experiments}
\subsection{Experimental settings}
\subsubsection{Dataset}

To evaluate our proposed method, multiple real underwater image datasets are utilized.
The UEQAB dataset~\cite{li2022human}, comprising 8000 images obtained through various image enhancement techniques, with each image associated with a MOS score, is utilized. In this dataset, 7200 images are randomly selected as the training set, while the remaining 800 images are utilized for testing, enabling the training and evaluation of our CLIP perception model. For training and testing the UIE network, the UIEB~\cite{li2019underwater}, U45~\cite{li1906fusion}, and SQUID~\cite{akkaynak2019sea} datasets are employed. Specifically, the UIEB dataset consists of 890 pairs of reference images and 60 non-reference challenging images. Following the settings of the paper~\cite{li2019underwater}, 800 reference images are randomly selected for training, and the remaining 90 reference images (U90) and 60 challenging images (C60) are used as the test set. The U45 test set consists of 45 real underwater images exhibiting color and contrast distortions. Furthermore, the SQUID dataset contains 57 pairs of images captured from four distinct diving locations, and we select 16 representative examples as the test set, which is the same as \cite{li2021underwater}.

\subsubsection{Implementation Details}
All experiments were conducted using the PyTorch framework on Ubuntu 20 with NVIDIA TITAN RTX for the CLIP perception model and UIE model. For the CLIP perception model, we utilized the SGD optimizer, with a fixed learning rate of 0.002, a batch size of 64, and a training iteration of 100,000. In our UIE model, the network structure from NU2Net~\cite{guo2023underwater} is leveraged, and integrated with the proposed perception loss and curriculum contrastive regularization. For the UIE model training, the batch size and the number of epochs were set as 16 and 800, respectively. The Adam optimizer with an initial learning rate of 0.001 is employed, and the cosine annealing strategy is utilized for adjusting the learning rate. For fair comparisons, all deep learning-based UIE methods were trained on the same devices and datasets. In our experiment, the input images were uniformly cropped to a size of $256\times 256$, and data augmentation was applied using random flipping. The non-easy negatives used were generated by the UDCP~\cite{drews2013transmission}, IBLA~\cite{peng2017underwater}, DCP~\cite{he2010single}, HE~\cite{hummel1975image}, FUnIE~\cite{islam2020fast}, and USUIR~\cite{fu2022unsupervised} methods. We followed the method  ~\cite{zheng2023curricular}, which extracted latent features extracted from the 1st, 3rd, 5th, 9th, and 13th layers of the fixed pre-trained VGG-19 model for $\mathcal{L}_1$ distance computation and set weights $\xi _{i} (i=1,2,3,4,5)$  to $1/32,1/16,1/8,1/4,1$ in Eq.(\ref{eq7}). The weighting parameters for $\mathcal{L}_{total}$ are as follows: $\lambda _1=0.025, \lambda _2=0.1$.

\subsubsection{Evaluation metrics}
For the test dataset U90 with reference images, Peak Signal-to-Noise Ratio (PSNR), Structure Similarity Index (SSIM), and Learned Perceptual Image Patch Similarity (LPIPS)~\cite{zhang2018unreasonable} are used as the image quality assessment metrics. These metrics represent the similarity between the enhanced images and reference images. A higher PSNR value indicates a closer resemblance in pixel fidelity between the two images, a higher SSIM value reflects more similar structure and texture while lower LPIPS indicates a higher extent of similarity between two images. 
For the test datasets C60, U45, and SQUID, which do not contain reference images, in addition to the two conventional non-reference evaluation metrics for Underwater images, i.e., Underwater Color Image Quality Evaluation (UCIQE)~\cite{yang2015underwater} and Underwater Image Quality Measurement (UIQM)~\cite{panetta2015human}, the proposed CLIP perception model is also employed for its stronger linear correlation with human visual perception. A higher score in UCIQE or UIQM indicates better color balance, sharpness, and contrast in the image while a higher CLIP perception score denotes better human perceptual quality.

\subsection{Quantitative Comparisons}
Eleven methods, including traditional methods and deep learning methods, are used as the comparison UIE methods. The traditional methods include UDCP~\cite{drews2013transmission}, IBLA~\cite{peng2017underwater}, MLLE~\cite{zhang2022underwater} and the deep learning methods include WaterNet~\cite{li2019underwater}, FUnIE~\cite{islam2020fast}, Shallow-UWnet~\cite{naik2021shallow}, Ucolor~\cite{li2021underwater}, PUIE-Net \cite{PUIE-Net}, Ushape~\cite{peng2023u}, PUGAN~\cite{10155564} and the NU2Net~\cite{guo2023underwater}. It can be observed from Table \ref{tab:table2} that traditional methods like UDCP, IBLA, and MLLE could not achieve good results because the physical models they employ cannot adequately capture the complex and diverse patterns present in underwater images. Some early deep learning-based UIE methods such as FUnIE, and Shallow-UWnet utilizing simplistic or shallow networks, obtain unsatisfactory results in UIE performance. In contrast, despite its simplicity, WaterNet yields superior performance by leveraging not only the original distorted input but also additional inputs from various preprocessing such as white balance and gamma correction.
By utilizing complicated network structures, PUGAN, PUIE-Net, and Ushape, could achieve further good UIE performance.
By comparison, NU2Net, incorporating its proposed URanker model for underwater images into the loss function, achieves the second-best results in terms of PSNR, SSIM. Compared with the NU2Net method, our proposed method exhibits improvements of 0.446 dB and 0.005 in PSNR and SSIM metrics, respectively, and a decrease of 0.021 in the LPIPS metric (lower is better). 
Given that our network structure is based on NU2Net, the observed improvements suggest that the incorporation of the proposed CLIP perception loss and curriculum contrastive regularization, which integrate the CLIP perception model, can effectively restore the structural details of the distorted image, resulting in enhanced perceptual quality.

\begin{table}[!t]
\caption{Quantitative comparison of the proposed CLIP perception model with other image quality assessment methods on the UEQAB test set. Bold font denotes the best results, while underlining indicates the second-best results.\label{tab:table1}}
\definecolor{Gray}{gray}{0.9}
\centering
\renewcommand\arraystretch{1.1}
\begin{tabular}{l|c|c}
\toprule
\textbf{Methods}                                & \textbf{PLCC$\uparrow$}         & \textbf{SROCC$\uparrow$}       \\
\midrule
UCIQE~\cite{yang2015underwater}                 & 0.35                  & 0.28      \\
UIQM~\cite{panetta2015human}                    & 0.44                  & 0.39     \\
NUIQ~\cite{jiang2022underwater}                 & 0.62                  & 0.65   \\
BFEN~\cite{wu2020subjective}                    & \underline{0.81}                  & \textbf{0.81}  \\
\midrule
\rowcolor{Gray}
\textbf{Ours}                                            &\textbf{0.83}   & \underline{0.80}                        \\
\bottomrule
\end{tabular}
\end{table}

\begin{table}[!t]
\caption{Quantitative comparison between different UIE methods on U90 test set. Bold font denotes the best results, while underlining indicates the second-best results.\label{tab:table2}}
\centering
\renewcommand\arraystretch{1.1}
\begin{tabular}{l|c|c|c}
\toprule
\textbf{Methods}      & \textbf{PSNR$\uparrow$} & \textbf{SSIM$\uparrow$}  & \textbf{LPIPS$\downarrow$} \\
\midrule
UDCP (ICCVW'13)       & 10.277        & 0.486               & 0.392          \\
IBLA (TIP'17)         & 15.046        & 0.683               & 0.316          \\
WaterNet (TIP'19)      & 20.998        & 0.919               & \underline{0.149}          \\
FUnIE (RAL'20)        & 19.454        & 0.871               & 0.175          \\
Shallow-UWnet (AAAI'21)        &18.120         &0.721        &0.289
\\
Ucolor (TIP'21)       & 20.730        & 0.900                 & 0.165          \\
MLLE (TIP'22)         & 18.977        & 0.841               & 0.275          \\
PUIE-Net (ECCV'22)    & 21.970        & 0.890               & 0.155   \\
Ushape (TIP'23)       & 20.920         & 0.853               & 0.206          \\
PUGAN (TIP'23)        &22.576         &0.920                &0.159
\\
NU2Net (AAAI'23, Oral) & \underline{22.669}        & \underline{0.924}               & 0.154          \\
\midrule
\rowcolor{Gray} \textbf{Ours}                & \textbf{23.115}              & \textbf{0.929}          & \textbf{0.133}          \\
\bottomrule 
\end{tabular}
\end{table}

\begin{table*}[]
\caption{Quantitative comparison of different UIE methods for the non-reference test set U45, SQUID, and C60. The top three results are marked with \textcolor{red}{{red}}, \textcolor{blue}{{blue}}, and \textcolor{green}{{green}}, respectively. \label{tab:table3}}
\centering
\renewcommand\arraystretch{1.14}
\scalebox{1}{\begin{tabular}{l|ccc|ccc|ccc}
\toprule
\multicolumn{1}{l|}{}                         & \multicolumn{3}{c|}{\textbf{U45}}                                        & \multicolumn{3}{c|}{\textbf{SQUID}}                                        & \multicolumn{3}{c}{\textbf{C60}}                                      \\ \cline{2-10} 
\multicolumn{1}{l|}{\multirow{-2}{*}{\textbf{Methods}}} & \multicolumn{1}{c}{\textbf{UCIQE$\uparrow$}} & \multicolumn{1}{c}{\textbf{UIQM$\uparrow$}} & \multicolumn{1}{c|}{\textbf{CLIP-Score$\uparrow$}} & \multicolumn{1}{c}{\textbf{UCIQE$\uparrow$}} & \multicolumn{1}{c}{\textbf{UIQM$\uparrow$}} & \multicolumn{1}{c|}{\textbf{CLIP-Score$\uparrow$}} & \multicolumn{1}{c}{\textbf{UCIQE$\uparrow$}} & \multicolumn{1}{c}{\textbf{UIQM$\uparrow$}}  & \multicolumn{1}{c}{\textbf{CLIP-Score$\uparrow$}}\\ \midrule
UDCP (ICCVW'13)         & 0.584          & 2.086          & 36.63          & 0.554          & 1.082          & 33.61          & 0.515          & 1.215       & 32.21          \\
IBLA (TIP'17)           & 0.579          & 1.672          & 40.81          & 0.466          & 0.866          & 46.73          & 0.564          & 1.893       & 37.66          \\
WaterNet (TIP'19)       & 0.582          & 3.295          & 54.41          & \textcolor{red}{{0.571}} & \textcolor{blue}{{2.518}}    & 51.98          & 0.566          & 2.653       & 50.28          \\
FUnIE (RAL'20)          & \textcolor{blue}{{0.599}}    & \textcolor{red}{{3.398}} & 48.43          & 0.532          & \textcolor{red}{{2.746}} & 52.03          & 0.570           & \textcolor{red}{{3.258}}       & 46.44          \\
Shallow-UWnet (AAAI'21) & 0.471          & 3.033          & 51.71         & 0.421          & 2.094           & 55.87           & 0.466          & 2.396       & 46.78        \\
Ucolor (TIP'21)         & 0.564          & 3.351          & \textcolor{green}{{55.91}}          & 0.514          & 2.215          & 57.86          & 0.532          & 2.746       & 50.23          \\
MLLE (TIP'22)           
& 0.598          & 2.599          & 55.01          
& 0.562          & 2.314          & 43.68          
& \textcolor{blue}{{0.581}}    & 2.310        & 49.49          \\

PUIE-Net (ECCV'22)    
& 0.578        & 3.199               & 54.56             
& 0.522        & 2.323               & \textcolor{blue}{{62.23}}
& 0.558        & 2.521               & \textcolor{red}{{51.45}}      \\

Ushape (TIP'23)         
& 0.553          & 3.248          & 55.01          
& 0.528          & 2.256          & 60.25          
& 0.534          & 2.783          & \textcolor{red}{{51.45}}                 \\

PUGAN (TIP'23)          
& \textcolor{blue}{{0.599}}         & 3.395          & 53.88          
& \textcolor{green}{{0.566}}        & 2.399          & 59.77         
& \textcolor{red}{{0.612}}          & \textcolor{blue}{{3.001}}  & 50.35        \\

NU2Net (AAAI’23, Oral)   
& 0.595          
& \textcolor{blue}{{3.396}}          
& \textcolor{red}{{56.89}} 
& 0.551          
& \textcolor{green}{{2.480}}          
& \textcolor{green}{{61.22}}   
& 0.564          
& \textcolor{green}{{2.900}}       
& \textcolor{blue}{{50.81}}    \\    

\midrule
\rowcolor{Gray}
\textbf{Ours}          
& \textcolor{red}{{0.601}}   
& \textcolor{red}{{3.398}} 
& \textcolor{blue}{{56.02}}    
& \textcolor{blue}{{0.570}}    & 2.360          & \textcolor{red}{{63.02}} 
& \textcolor{green}{{0.573}}   & 2.810          & \textcolor{green}{{50.41}}   \\
\bottomrule
\end{tabular}}
\end{table*}

In addition, to validate the robustness of our model, the performance of different methods on the non-reference datasets is also evaluated. The higher UIQM and UCIQE metrics indicate that the enhanced images have better contrast and sharp colors. But according to previous research ~\cite{li2019underwater}, UCIQE and UIQM are biased against certain features rather than evaluating the entire image, neglecting considerations for color shifts and artifacts. By contrast, the CLIP perceptual score could better reflect the overall quality of enhanced underwater images. As shown in Table \ref{tab:table3}, it can be found that traditional methods such as UDCP and MLLE tend to achieve higher UCIQE scores, and the deep learning-based method FUnIE could achieve the highest UIQM scores across three datasets. However, as illustrated in Fig. \ref{fig:No-reference}, MLLE tends to introduce obvious chromatic aberrations on enhanced images while FUnIE generates local patch artifacts particularly obvious on example images from SQUID. These findings align with the conclusion drawn in ~\cite{li2019underwater} and the lower performance in Table \ref{tab:table1}. It can be observed that only NU2Net and the proposed method could achieve both high scores in  UIQM and UCIQE metrics as well as CLIP perception model metrics. In contrast, our proposed method enhances the overall perceptual quality of underwater images while preserving local details, resulting in improved contrast and vibrant colors.


\subsection{Visual Comparisons}

\begin{figure*}
\centering
\includegraphics[width=1\linewidth]{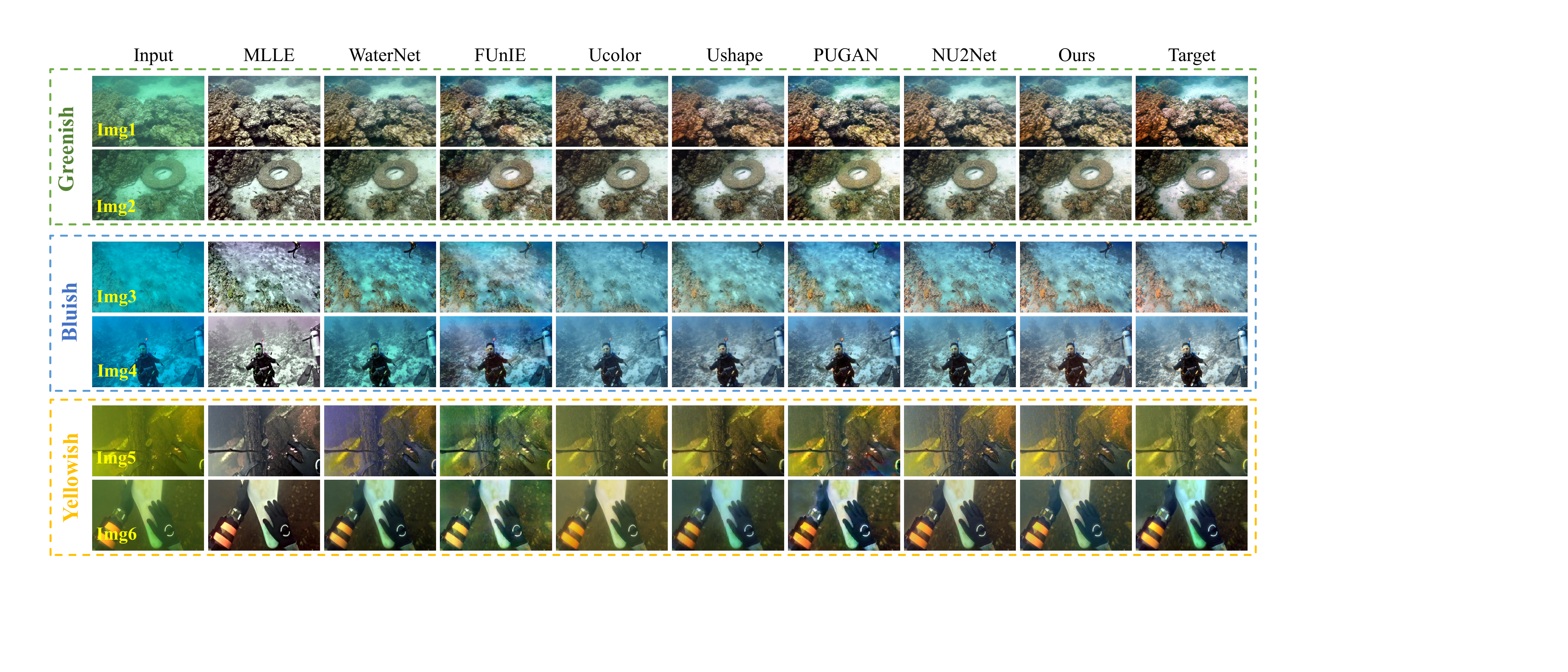}
\caption{\label{fig:U90}Visual comparison of enhancement results for the U90 test set. From left to right, the original underwater image, MLLE~\cite{zhang2022underwater}, WaterNet~\cite{li2019underwater}, FUnIE~\cite{islam2020fast}, Ucolor~\cite{li2021underwater}, Ushape~\cite{peng2023u}, PUGAN~\cite{10155564}, NU2Net~\cite{guo2023underwater}, our method, and the reference image.}
\vspace{-0.99 em}  
\end{figure*}

In this section, the visual results of different test sets are comprehensively compared. Firstly, the enhancement effect of different methods on the U90 dataset is compared, as shown in Fig. \ref{fig:U90}. It can be observed that the input images are suffering from color distortions such as greenish, bluish, or yellowish, as well as low contrast and blurriness. Traditional methods like MLLE sometimes introduce additional chromatic aberrations and cannot effectively handle image color distortions. Deep learning methods are clearly outperforming traditional methods but still have some limitations. The WaterNet method struggles to effectively address images with blue color distortion and tends to darken the brightness of the enhanced images. The FUnIE method may still be introducing some local patch artifacts in the repaired images. The Ucolor and Ushape methods have a poor perception of texture details, and the texture details of the enhanced images are easily missing. In addition, the enhanced images from the NU2Net network are sometimes showing chromatic aberration problems such as a bluish color, indicating that the color of the enhanced images is not being effectively controlled. It can be observed that our proposed method effectively alleviates this issue. The enhanced images display less color deviation, along with improved contrast and structural details.
This can be primarily attributed to the utilization of CLIP perception loss and curriculum contrastive regularization in our approach. The CLIP perception loss forces the model to converge towards better perceptual results, while the curriculum contrastive regularization enables accurate control of the color enhancement in the enhanced images to avoid under- or over-enhancement.

\begin{figure*}
\centering
\includegraphics[width=1\linewidth]{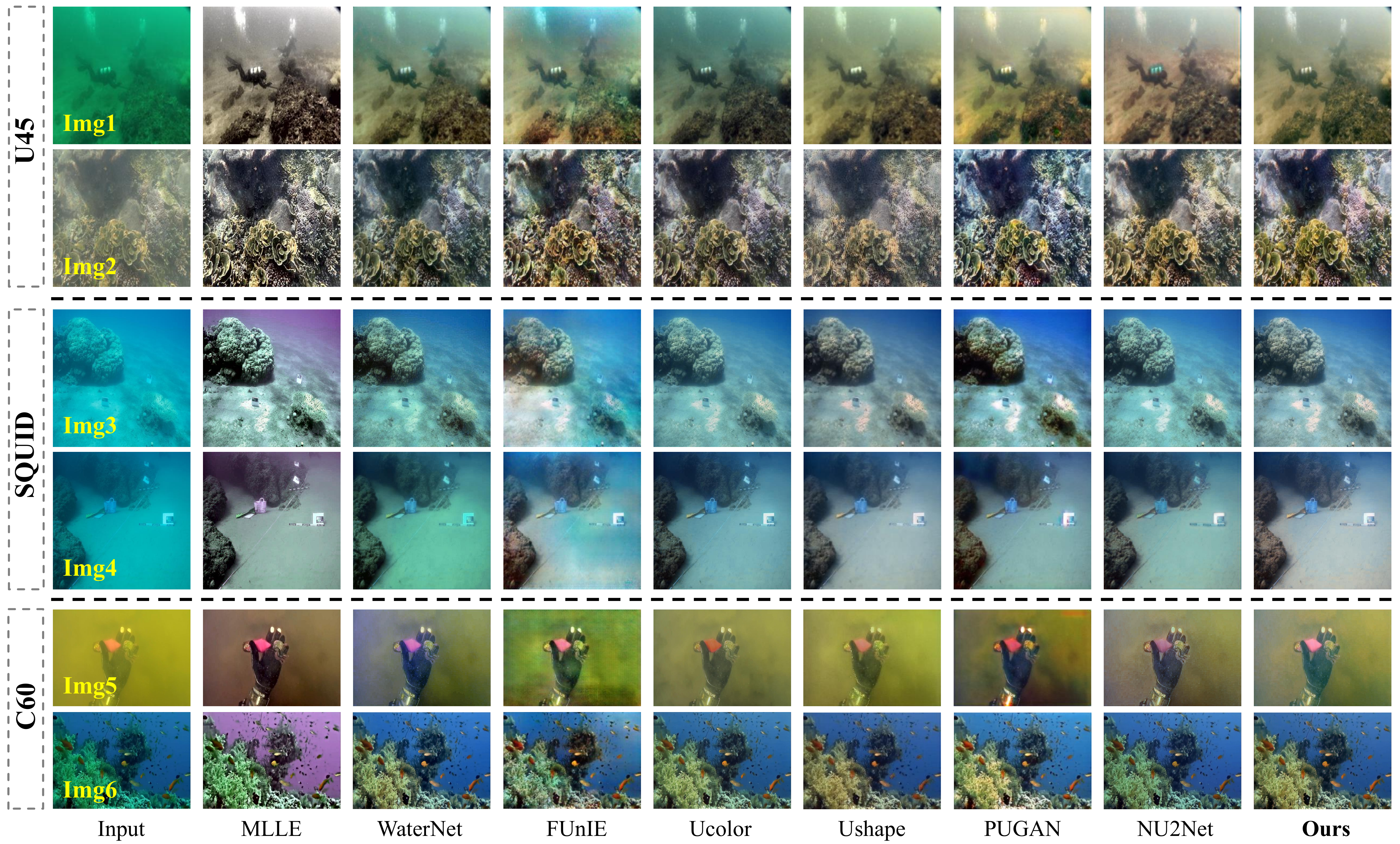}
\caption{\label{fig:No-reference}Visual comparison of enhancement results for the U45, SQUID, and C60 test set. From left to right, the original underwater image, MLLE~\cite{zhang2022underwater}, WaterNet~\cite{li2019underwater}, FUnIE~\cite{islam2020fast}, Ucolor~\cite{li2021underwater}, Ushape~\cite{peng2023u}, PUGAN~\cite{10155564}, NU2Net~\cite{guo2023underwater} and our method.}
\end{figure*}

Visual comparisons are also conducted across three no-reference underwater datasets, i.e., U45 dataset, SQUID dataset, and C60 dataset. The U45 dataset includes various types of distortions such as bluish, greenish, and low-contrast. The SUIQD dataset represents different underwater environments. Meanwhile, the C60 dataset exhibits more severe color distortion and blurriness. As shown in Fig. \ref{fig:No-reference}. Many traditional and deep learning methods do not perform well on these datasets because they contain a diverse underwater environment with complex and variable types of distortion. For instance, the enhanced images produced by the MLLE method frequently introduce additional color distortions. Images enhanced by methods like WaterNet and FUnIE still show evident color distortions. Although the Ushape method has advanced in enhancement efficacy beyond these techniques, it has its own limitations. It performs poorly on the blurry, distorted images from the U45 dataset, and the images it enhances from the C60 dataset sometimes retain a blue color bias. Similarly, the NU2Net method yields unsatisfactory enhancement results on both the U45 and SQUID datasets.
Clearly, when the image distortion is significant, these methods struggle to simultaneously address color distortion, structural distortion, and detail blurring issues. In contrast, it can be observed that our method generally exhibits better visual quality across the three datasets, i.e., the enhanced image scarcely exhibits color distortion or introduces additional chromatic aberration. The visual results further indicate that our method has superior robustness and generalization capability.


\subsection{Individual roles of Modules in the Proposed Loss Function}
\begin{figure}
\centering
\includegraphics[width=1\linewidth]{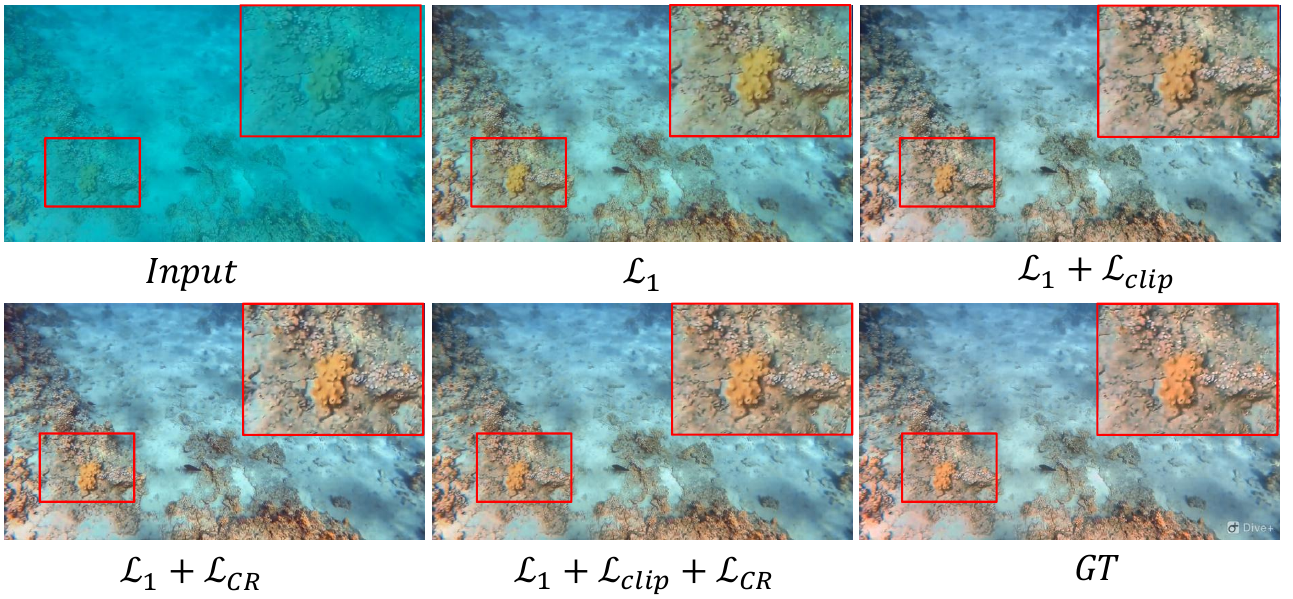}
\caption{\label{fig:Ablation_1}Visual effects of enhanced images obtained by schemes with individual loss
modules and the proposed scheme.}
\end{figure}

To explore the individual role of the perception loss and curriculum contrastive regularization with the CLIP model in the overall loss function Eq.\ref{eq8}, ablation studies have been conducted. 
 Additionally, to delve deeper into the role of the CLIP perception model in identifying non-easy negatives in the regularization module, PSNR is employed instead of the CLIP model, denoted as $\mathcal{L}_{CR(PSNR)}$. 
As depicted in Table \ref{tab:table4}, when setting the loss function with only $\mathcal{L}_1$ as the baseline, it can be observed that the individual modules of $\mathcal{L}_{clip}$ and $\mathcal{L}_{CR(PSNR)}$ elevate PSNR performance by 0.258 and 0.632 respectively, and SSIM performance by 0.002 and 0.007 respectively. Furthermore, the joint integration of $\mathcal{L}_{clip}$ and $\mathcal{L}_{CR(PSNR)}$ enhances the baseline performance with a PSNR gain of 0.696. Additionally, it could be found that the scheme with $\mathcal{L}_1+\mathcal{L}_{clip}+\mathcal{L}_{CR(PSNR)}$ is inferior to that with $\mathcal{L}_1+\mathcal{L}_{clip}+\mathcal{L}_{CR}$, indicating the appreciable role of the CLIP model in the regularization module. Fig. \ref{fig:Ablation_1} provides the visual effects of different schemes with individual loss modules and the proposed scheme. It can be observed that with the successive addition of the proposed loss modules, the enhanced underwater image increasingly resembles the ground truth image.

\subsection{Effects of Quality Assessment Model on the Proposed Perception Loss}
To test the efficacy of the proposed perception loss module with the CLIP perception model, two quality assessment models, i.e., Uranker~\cite{guo2023underwater}  and  BFEN~\cite{wu2020subjective}, were adopted and incorporated as the perceptual loss as $\mathcal{L}_{Uranker}$ and $\mathcal{L}_{BFEN}$, respectively. In this study, the scheme  $\mathcal{L}_1+\mathcal{L}_{clip}$ is regarded as the anchor. Note the scheme $\mathcal{L}_1+\mathcal{L}_{clip}$ actually means NU2Net method ~\cite{guo2023underwater}. From Table \ref{tab:table5}, it can be observed that the scheme $\mathcal{L}_1+\mathcal{L}_{Uranker}$ is superior to the two other schemes incorporating two other quality assessment models for underwater images, i.e., $\mathcal{L}_1+\mathcal{L}_{Uranker}$ and $\mathcal{L}_1+\mathcal{L}_{BFEN}$. This fact validates the effectiveness of the proposed perception loss module with the CLIP perception model.

\subsection{Effects of negative number in the Proposed Contrastive Regularization module}
The effectiveness of the curriculum contrastive regularization module mainly results from the constraints on the solution space by multiple negatives. To explore the impact of the number of negatives used in Contrastive Regularization (CR) module on the UIE performance, two other schemes configured with different numbers of negatives were conducted. One adopts only one simple negative; the other adopts nine negatives,  including two extra non-easy negatives generated using the Ushape and PUIE~\cite{fu2022uncertainty} methods, in addition to the seven negatives generated using the same methods employed in the proposed regularization module. The two schemes and the proposed scheme are listed as ``+CR(1:1)'', ``+CR(1:9)'', and ``+CR(1:7)'', respectively while the proposed model without $\mathcal{L}_{CR}$ in the loss function, namely with only  $\mathcal{L}_1+\mathcal{L}_{Uranker}$, lists as the anchor.  From Table \ref{tab:table6}, as the number of the negatives used in CR increases from 1 to 9, the performance first increases and then decreases, and ``+CR(1:7)'' achieves the best performance. The reason could be that more negatives in CR could provide effective feature constraints on the solution space than only one single negative. However, as the number of negatives continues to increase, it may interfere with model training due to the uncertainty the non-easy negatives bring. In our experiment, the number of negatives is set as 7.

\subsection{Effects of the Weights between CLIP Perception Loss and Regularization Loss}
To investigate the impact of weights $\lambda_1$ and $\lambda_2$ in Eq. \ref{eq8} on the overall UIE performance, related experiments have been conducted by sweeping one parameter and fixing the other. As demonstrated in Table \ref{tab:table8}, when fixing $\lambda_2$ and varying $\lambda_1$, peak performance is observed around $\lambda_1 = 0.025$. Similarly, when fixing $\lambda_1$ and adjusting $\lambda_2$, the best performance occurs around $\lambda_2 = 0.1$. Thus, in our method,  $\lambda_1$ and  $\lambda_2$ are set as 0.025 and 0.1, respectively, which implies that the regularization loss plays a more significant role than the CLIP perception loss.


\subsection{Discussion on the baseline networks}
\begin{figure}[!t]
\centering
\includegraphics[width=1\linewidth]{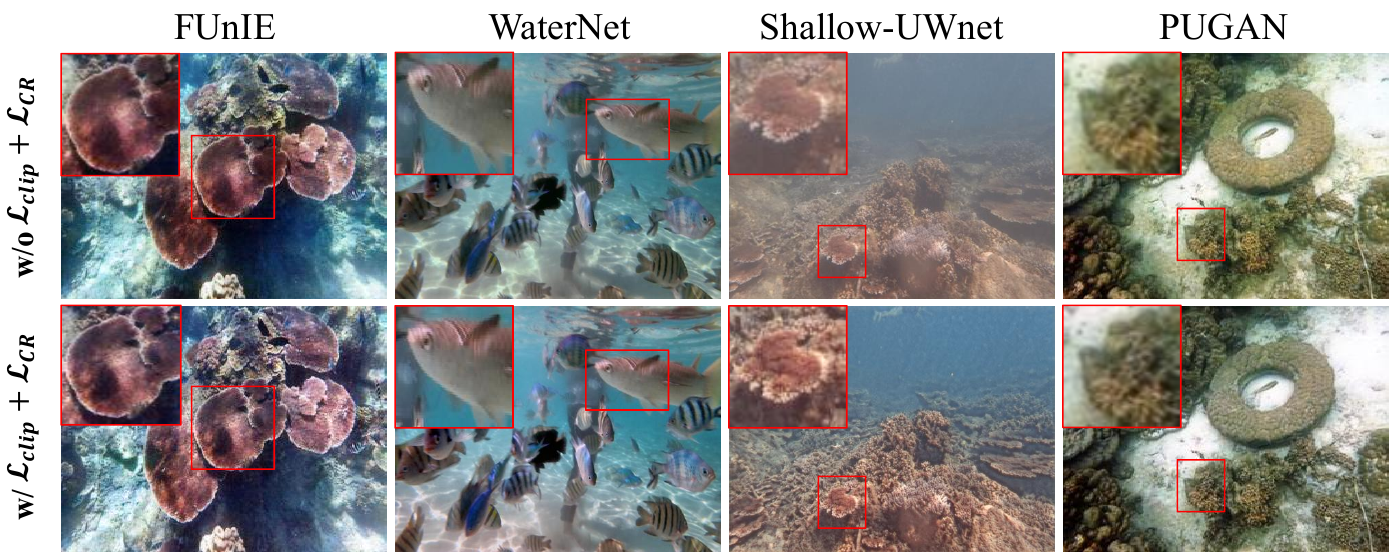}
\caption{\label{fig:Ablation}Visual comparison between four UIE models without and with $\mathcal{L}_{clip}+\mathcal{L}_{CR}$.}
\end{figure}

To further validate the effectiveness of the proposed loss modules, i.e., the perception loss and curriculum contrastive regularization with the CLIP model, several UIE networks beyond NU2Net are considered, including WaterNet ~\cite{li2019underwater}, FUnIE~\cite{islam2020fast}, Shallow-UWnet~\cite{naik2021shallow}, and PUGAN~\cite{10155564}, as baseline networks.
In Table \ref{tab:table7}, the column w/o $\mathcal{L}_{clip}+\mathcal{L}_{CR}$ lists the results of these UIE models with their original loss. The column w $\mathcal{L}_{clip}+\mathcal{L}_{CR}$ indicates  
integrating the $\mathcal{L}_{clip}$ and $\mathcal{L}_{CR}$ into these UIE networks as additional loss. Note that, since FUnIE and Shallow-UWnet employ VGG loss \cite{johnson2016perceptual} in their loss functions which overlaps with the feature extraction in the proposed curriculum contrastive regularization, the VGG loss is removed when adding $\mathcal{L}_{clip}+\mathcal{L}_{CR}$. It can be observed that the performance of all four UIE networks improves when $\mathcal{L}_{clip}$ and $\mathcal{L}_{CR}$ are added as additional losses. Notably, with Shallow-UWnet as the baseline network, the PSNR gain could reach 1.031.
Fig. \ref{fig:Ablation} demonstrates the visual comparison between the four UIE models without and with $\mathcal{L}_{clip}+\mathcal{L}_{CR}$. It is evident that the integration of the proposed perception loss and regularization module has guided the four UIE models to generate enhanced underwater images with superior visual quality, featuring improved contrast, sharper edges, and finer details.

\begin{table}[]
\caption{The experimental results of schemes with different combinations of loss modules.\label{tab:table4}}
\centering
\renewcommand\arraystretch{1.1}
\begin{tabular}{cccc|c|c}
\toprule
$\mathcal{L}_1$  &$\mathcal{L}_{clip}$  &$\mathcal{L}_{CR}$  &$\mathcal{L}_{CR(PSNR)}$  & PSNR$\uparrow$   & SSIM$\uparrow$  \\
\midrule
$\checkmark$  &    &    &                          & 22.419 & 0.922     \\
$\checkmark$  & $\checkmark$ &    &                & 22.677 & 0.924     \\
$\checkmark$  &   & $\checkmark$  &                & 23.051 & 0.929     \\
$\checkmark$  & $\checkmark$  &    & $\checkmark$  & 23.078 & 0.929    \\
$\checkmark$  & $\checkmark$  &$\checkmark$   &    & \textbf{23.115} & \textbf{0.929}  \\
\bottomrule
\end{tabular}
\end{table}

\begin{table}[!t]
\caption{Effects of quality assessment models incorporated as the perception loss module.\label{tab:table5}}
\centering
\renewcommand\arraystretch{1.1}
\begin{tabular}{l|c|c}
\toprule
Loss Module & PSNR$\uparrow$   & SSIM$\uparrow$  \\
\midrule
$\mathcal{L}_1+\mathcal{L}_{Uranker}$                    & 22.669 & 0.924 \\
$\mathcal{L}_1+\mathcal{L}_{BFEN}$                      & 22.620  & 0.921 \\
$\mathcal{L}_1+\mathcal{L}_{clip}$                       & \textbf{22.677} & \textbf{0.924} \\
\bottomrule
\end{tabular}
\end{table}

\begin{table}[!t]
\caption{Effects of the number of negatives on the regularization module.\label{tab:table6}}
\centering
\renewcommand\arraystretch{1.2}
\begin{tabular}{c|c|c}
\toprule
Number of negatives & PSNR$\uparrow$   & SSIM$\uparrow$  \\
\midrule
Baseline (w/o $\mathcal{L}_{CR}$)  & 22.677   & 0.924  \\  
+ CR(easy)           & 22.949   & 0.923  \\
+ CR(1:9)           & 22.979   & 0.929 \\
+ CR(1:7) (\textbf{Ours})            & \textbf{23.115}    & \textbf{0.929} \\
\bottomrule
\end{tabular}
\end{table}

\begin{table}[!t]
\caption{Effects of weight values of $\lambda _1$ and $\lambda _2$ in the proposed loss function.\label{tab:table8}}

\centering
\renewcommand\arraystretch{1.1}
\begin{tabular}{c|c|c|c}
\toprule
\textbf{$\lambda _1$} & \textbf{$\lambda _2$} & {PSNR$\uparrow$} & {SSIM$\uparrow$} \\
\midrule
0.025      & 0.025      & 22.850         & 0.930         \\
0.025      & 0.055      & 22.896        & 0.929         \\
\textbf{0.025}      & \textbf{0.100}      & \textbf{23.115}        & \textbf{0.929}         \\
0.025     &0.150       & 22.995            &0.925  \\
0.020      & 0.100      & 22.870        & 0.929         \\
0.050      & 0.100      & 22.810        & 0.928         \\
0.100      & 0.100      & 22.812        & 0.928     \\
\bottomrule
\end{tabular}
\end{table}

\begin{table}[!t]
\caption{Performance comparison on U90 test set between different UIE models without and with adding the $\mathcal{L}_{clip}$ and $\mathcal{L}_{CR}$ .\label{tab:table7}}
\centering
\renewcommand\arraystretch{1.2}
 \resizebox{\linewidth}{!}{\begin{tabular}{l|cc|cc}
\toprule
\multicolumn{1}{l|}{\multirow{3}{*}{Methods}} & \multicolumn{2}{c|}{\multirow{2}{*}{w/o $\mathcal{L}_{clip}+\mathcal{L}_{CR}$}}             & \multicolumn{2}{c}{\multirow{2}{*}{w/ $\mathcal{L}_{clip}+\mathcal{L}_{CR}$}}             \\
\multicolumn{1}{c|}{}                         & \multicolumn{2}{c|}{}                                 & \multicolumn{2}{c}{}                                 \\ \cline{2-5} 
\multicolumn{1}{c|}{}                         & \multicolumn{1}{c}{PSNR$\uparrow$} & \multicolumn{1}{c|}{SSIM$\uparrow$} & \multicolumn{1}{c}{PSNR$\uparrow$({\color[HTML]{FF0000}$\Delta$})} & \multicolumn{1}{c}{SSIM$\uparrow$({\color[HTML]{FF0000}$\Delta$})} \\ \midrule
WaterNet~\cite{li2019underwater}                                        & 20.998                    & 0.919                     & 21.324({\color[HTML]{FF0000}0.326})                    & 0.927({\color[HTML]{FF0000}0.008})                     \\
FUnIE~\cite{islam2020fast}                                          & 19.454                    & 0.871                     & 19.711({\color[HTML]{FF0000}0.257})                    & 0.891({\color[HTML]{FF0000}0.019})                  \\
Shallow-UWnet~\cite{naik2021shallow}                                   &18.120                    &0.721                       &19.151({\color[HTML]{FF0000}1.031})                     &0.801({\color[HTML]{FF0000}0.080})                 \\
PUGAN~\cite{10155564}  &22.576      &0.920       &22.811({\color[HTML]{FF0000}0.235})   &0.921({\color[HTML]{FF0000}0.001})   \\
NU2Net~\cite{guo2023underwater}  &22.669      &0.924       &23.115({\color[HTML]{FF0000}0.446})   &0.929({\color[HTML]{FF0000}0.005})   \\
\bottomrule
\end{tabular}}
\end{table}

\begin{figure}[!t]
\centering
\includegraphics[width=1\linewidth]{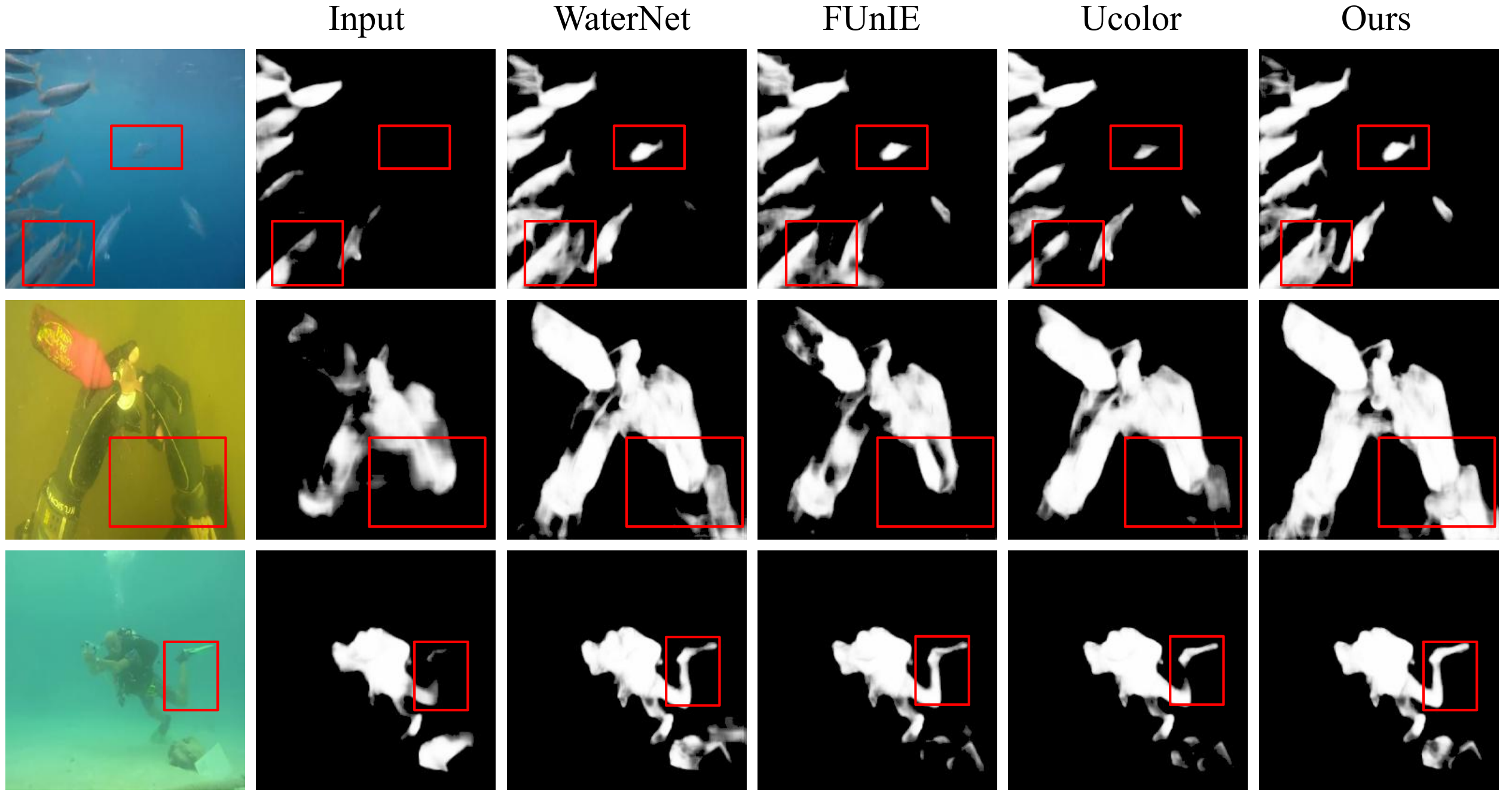}
\caption{\label{fig:svam}The saliency detection results by SVAM-Net~\cite{2020SVAM} on the original input and enhanced underwater images by WaterNet, FUnIE, Ucolor, and the proposed method, respectively.
}
\end{figure}

\subsection{Application Test to High-level Task}
To explore the applicability of the proposed UIE method in real scenarios, the saliency object detection model SVAM-Net~\cite{2020SVAM} is used to perform target detection on the original input images and the associated images enhanced by several comparative UIE methods and the proposed method. 
Fig. \ref{fig:svam} shows the object detection results for various underwater scenes with different color tones such as blue, yellow, and green. It can be observed that the target objects in the original input image and the enhanced images obtained by the comparative UIE methods are difficult to be fully detected. However, after applying our proposed UIE method, the target objects in enhanced images can be generally detected, significantly improving the accuracy of the target detection task. This demonstrates the practicality of our proposed UIE method in object detection and recognition.

\subsection{Limitation and Discussion}
The effectiveness and excellent performance of our method in underwater image enhancement tasks is demonstrated by conducting experiments on multiple datasets. However, there are limitations inherent in our method due to the lack of sufficient training datasets and the incomplete optimization of the CLIP perception model. Specifically, the CLIP perception model may benefit from a more accurate selection of initialization prompts to enhance its accuracy in evaluating underwater image quality, particularly in terms of color distortion and texture structure. Although we have made initial progress in exploring the potential of CLIP in underwater image enhancement, how to better integrate CLIP models with existing enhancement models still worth further investigation and exploration. We believe that more relevant research will emerge in the future to make more progress in this field.

\section{Conclusion}
In this paper, we have
introduced an Under Water Image enhancement (UIE) framework in which the perception loss and curriculum contrastive regularization with CLIP model are proposed and incorporated into the UIE model. Specifically, the rich prior knowledge in the CLIP model is leveraged by employing prompt learning to train a CLIP perception model for underwater images. Subsequently, this perception model is incorporated as a perception loss module into the UIE model, aiming to enhance underwater images with better human perceptual quality.
Furthermore, the curriculum contrastive regularization utilizing multiple negatives is proposed to impose on the enhanced images within the CLIP perceptual space, in which the difficulty levels for negatives with diverse degrees of quality are measured. This approach effectively utilizes the latent features of different distorted images, preventing inadequate or excessive enhancement results. Extensive experiments validate the superiority and generalizability of the proposed UIE method, showcasing its effectiveness not only on underwater datasets with reference images but also on datasets containing non-reference images with diverse and complex distortions.




\bibliographystyle{IEEEtran}

\begin{thebibliography}{10}
\providecommand{\url}[1]{#1}
\csname url@samestyle\endcsname
\providecommand{\newblock}{\relax}
\providecommand{\bibinfo}[2]{#2}
\providecommand{\BIBentrySTDinterwordspacing}{\spaceskip=0pt\relax}
\providecommand{\BIBentryALTinterwordstretchfactor}{4}
\providecommand{\BIBentryALTinterwordspacing}{\spaceskip=\fontdimen2\font plus
\BIBentryALTinterwordstretchfactor\fontdimen3\font minus \fontdimen4\font\relax}
\providecommand{\BIBforeignlanguage}[2]{{%
\expandafter\ifx\csname l@#1\endcsname\relax
\typeout{** WARNING: IEEEtran.bst: No hyphenation pattern has been}%
\typeout{** loaded for the language `#1'. Using the pattern for}%
\typeout{** the default language instead.}%
\else
\language=\csname l@#1\endcsname
\fi
#2}}
\providecommand{\BIBdecl}{\relax}
\BIBdecl

\bibitem{PUIE-Net}
Z.~Fu, W.~Wang, Y.~Huang, X.~Ding, and K.-K. Ma, ``Uncertainty inspired underwater image enhancement,'' in \emph{European conference on computer vision}, 2022, pp. 465--482.

\bibitem{yang2019depth}
M.~Yang, J.~Hu, C.~Li, G.~Rohde, Y.~Du, and K.~Hu, ``An in-depth survey of underwater image enhancement and restoration,'' \emph{IEEE Access}, vol.~7, pp. 123\,638--123\,657, 2019.

\bibitem{2}
P.~Sahu, N.~Gupta, and N.~Sharma, ``A survey on underwater image enhancement techniques,'' \emph{International Journal of Computer Applications}, vol.~87, no.~13, 2014.

\bibitem{3}
M.~J. Kaiser \emph{et~al.}, \emph{Marine ecology: processes, systems, and impacts}.\hskip 1em plus 0.5em minus 0.4em\relax Oxford University Press, USA, 2011.

\bibitem{long2011marine}
R.~Long, ``The marine strategy framework directive: a new european approach to the regulation of the marine environment, marine natural resources and marine ecological services,'' \emph{Journal of Energy \& Natural Resources Law}, vol.~29, no.~1, pp. 1--44, 2011.

\bibitem{todd2019towards}
P.~A. Todd, E.~C. Heery, L.~H. Loke, R.~H. Thurstan, D.~J. Kotze, and C.~Swan, ``Towards an urban marine ecology: characterizing the drivers, patterns and processes of marine ecosystems in coastal cities,'' \emph{Oikos}, vol. 128, no.~9, pp. 1215--1242, 2019.

\bibitem{schettini2010underwater}
R.~Schettini and S.~Corchs, ``Underwater image processing: state of the art of restoration and image enhancement methods,'' \emph{EURASIP journal on advances in signal processing}, vol. 2010, pp. 1--14, 2010.

\bibitem{ancuti2012enhancing}
C.~Ancuti, C.~O. Ancuti, T.~Haber, and P.~Bekaert, ``Enhancing underwater images and videos by fusion,'' in \emph{2012 IEEE conference on computer vision and pattern recognition}.\hskip 1em plus 0.5em minus 0.4em\relax IEEE, 2012, pp. 81--88.

\bibitem{iqbal2010enhancing}
K.~Iqbal, M.~Odetayo, A.~James, R.~A. Salam, and A.~Z.~H. Talib, ``Enhancing the low quality images using unsupervised colour correction method,'' in \emph{2010 IEEE International Conference on Systems, Man and Cybernetics}.\hskip 1em plus 0.5em minus 0.4em\relax IEEE, 2010, pp. 1703--1709.

\bibitem{he2010single}
K.~He, J.~Sun, and X.~Tang, ``Single image haze removal using dark channel prior,'' \emph{IEEE transactions on pattern analysis and machine intelligence}, vol.~33, no.~12, pp. 2341--2353, 2010.

\bibitem{drews2016underwater}
P.~L. Drews, E.~R. Nascimento, S.~S. Botelho, and M.~F.~M. Campos, ``Underwater depth estimation and image restoration based on single images,'' \emph{IEEE computer graphics and applications}, vol.~36, no.~2, pp. 24--35, 2016.

\bibitem{li2016underwater}
C.-Y. Li, J.-C. Guo, R.-M. Cong, Y.-W. Pang, and B.~Wang, ``Underwater image enhancement by dehazing with minimum information loss and histogram distribution prior,'' \emph{IEEE Transactions on Image Processing}, vol.~25, no.~12, pp. 5664--5677, 2016.

\bibitem{zou2023object}
Z.~Zou, K.~Chen, Z.~Shi, Y.~Guo, and J.~Ye, ``Object detection in 20 years: A survey,'' \emph{Proceedings of the IEEE}, 2023.

\bibitem{guo2019underwater}
Y.~Guo, H.~Li, and P.~Zhuang, ``Underwater image enhancement using a multiscale dense generative adversarial network,'' \emph{IEEE Journal of Oceanic Engineering}, vol.~45, no.~3, pp. 862--870, 2019.

\bibitem{li2017watergan}
J.~Li, K.~A. Skinner, R.~M. Eustice, and M.~Johnson-Roberson, ``{WaterGAN}: Unsupervised generative network to enable real-time color correction of monocular underwater images,'' \emph{IEEE Robotics and Automation letters}, vol.~3, no.~1, pp. 387--394, 2017.

\bibitem{li2021underwater}
C.~Li, S.~Anwar, J.~Hou, R.~Cong, C.~Guo, and W.~Ren, ``Underwater image enhancement via medium transmission-guided multi-color space embedding,'' \emph{IEEE Transactions on Image Processing}, vol.~30, pp. 4985--5000, 2021.

\bibitem{akkaynak2019sea}
D.~Akkaynak and T.~Treibitz, ``Sea-thru: A method for removing water from underwater images,'' in \emph{Proceedings of the IEEE/CVF conference on computer vision and pattern recognition}, 2019, pp. 1682--1691.

\bibitem{li2019underwater}
C.~Li, C.~Guo, W.~Ren, R.~Cong, J.~Hou, S.~Kwong, and D.~Tao, ``An underwater image enhancement benchmark dataset and beyond,'' \emph{IEEE Transactions on Image Processing}, vol.~29, pp. 4376--4389, 2019.

\bibitem{peng2023u}
L.~Peng, C.~Zhu, and L.~Bian, ``U-shape transformer for underwater image enhancement,'' \emph{IEEE Transactions on Image Processing}, 2023.

\bibitem{guo2023underwater}
C.~Guo, R.~Wu, X.~Jin, L.~Han, W.~Zhang, Z.~Chai, and C.~Li, ``Underwater ranker: Learn which is better and how to be better,'' in \emph{Proceedings of the AAAI conference on artificial intelligence}, vol.~37, no.~1, 2023, pp. 702--709.

\bibitem{wang2023exploring}
J.~Wang, K.~C. Chan, and C.~C. Loy, ``Exploring clip for assessing the look and feel of images,'' in \emph{Proceedings of the AAAI Conference on Artificial Intelligence}, vol.~37, no.~2, 2023, pp. 2555--2563.

\bibitem{zheng2023curricular}
Y.~Zheng, J.~Zhan, S.~He, J.~Dong, and Y.~Du, ``Curricular contrastive regularization for physics-aware single image dehazing,'' in \emph{Proceedings of the IEEE/CVF Conference on Computer Vision and Pattern Recognition}, 2023, pp. 5785--5794.

\bibitem{radford2021learning}
A.~Radford, J.~W. Kim, C.~Hallacy, A.~Ramesh, G.~Goh, S.~Agarwal, G.~Sastry, A.~Askell, P.~Mishkin, J.~Clark \emph{et~al.}, ``Learning transferable visual models from natural language supervision,'' in \emph{International conference on machine learning}.\hskip 1em plus 0.5em minus 0.4em\relax PMLR, 2021, pp. 8748--8763.

\bibitem{wu2021contrastive}
H.~Wu, Y.~Qu, S.~Lin, J.~Zhou, R.~Qiao, Z.~Zhang, Y.~Xie, and L.~Ma, ``Contrastive learning for compact single image dehazing,'' in \emph{Proceedings of the IEEE/CVF Conference on Computer Vision and Pattern Recognition}, 2021, pp. 10\,551--10\,560.

\bibitem{peng2018generalization}
Y.-T. Peng, K.~Cao, and P.~C. Cosman, ``Generalization of the dark channel prior for single image restoration,'' \emph{IEEE Transactions on Image Processing}, vol.~27, no.~6, pp. 2856--2868, 2018.

\bibitem{galdran2015automatic}
A.~Galdran, D.~Pardo, A.~Pic{\'o}n, and A.~Alvarez-Gila, ``Automatic red-channel underwater image restoration,'' \emph{Journal of Visual Communication and Image Representation}, vol.~26, pp. 132--145, 2015.

\bibitem{peng2017underwater}
Y.-T. Peng and P.~C. Cosman, ``Underwater image restoration based on image blurriness and light absorption,'' \emph{IEEE transactions on image processing}, vol.~26, no.~4, pp. 1579--1594, 2017.

\bibitem{naik2021shallow}
A.~Naik, A.~Swarnakar, and K.~Mittal, ``Shallow-uwnet: Compressed model for underwater image enhancement (student abstract),'' in \emph{Proceedings of the AAAI Conference on Artificial Intelligence}, vol.~35, no.~18, 2021, pp. 15\,853--15\,854.

\bibitem{wang2021uiec}
Y.~Wang, J.~Guo, H.~Gao, and H.~Yue, ``{UIEC\^{} 2-Net}: Cnn-based underwater image enhancement using two color space,'' \emph{Signal Processing: Image Communication}, vol.~96, p. 116250, 2021.

\bibitem{fu2022uncertainty}
Z.~Fu, W.~Wang, Y.~Huang, X.~Ding, and K.-K. Ma, ``Uncertainty inspired underwater image enhancement,'' in \emph{European Conference on Computer Vision}.\hskip 1em plus 0.5em minus 0.4em\relax Springer, 2022, pp. 465--482.

\bibitem{fabbri2018enhancing}
C.~Fabbri, M.~J. Islam, and J.~Sattar, ``Enhancing underwater imagery using generative adversarial networks,'' in \emph{2018 IEEE international conference on robotics and automation (ICRA)}.\hskip 1em plus 0.5em minus 0.4em\relax IEEE, 2018, pp. 7159--7165.

\bibitem{zang2022open}
Y.~Zang, W.~Li, K.~Zhou, C.~Huang, and C.~C. Loy, ``Open-vocabulary detr with conditional matching,'' in \emph{European Conference on Computer Vision}.\hskip 1em plus 0.5em minus 0.4em\relax Springer, 2022, pp. 106--122.

\bibitem{kuo2022f}
W.~Kuo, Y.~Cui, X.~Gu, A.~Piergiovanni, and A.~Angelova, ``F-vlm: Open-vocabulary object detection upon frozen vision and language models,'' \emph{arXiv preprint arXiv:2209.15639}, 2022.

\bibitem{zhou2022extract}
C.~Zhou, C.~C. Loy, and B.~Dai, ``Extract free dense labels from clip,'' in \emph{European Conference on Computer Vision}.\hskip 1em plus 0.5em minus 0.4em\relax Springer, 2022, pp. 696--712.

\bibitem{zhang2023blind}
W.~Zhang, G.~Zhai, Y.~Wei, X.~Yang, and K.~Ma, ``Blind image quality assessment via vision-language correspondence: A multitask learning perspective,'' in \emph{Proceedings of the IEEE/CVF Conference on Computer Vision and Pattern Recognition}, 2023, pp. 14\,071--14\,081.

\bibitem{ke2023vila}
J.~Ke, K.~Ye, J.~Yu, Y.~Wu, P.~Milanfar, and F.~Yang, ``{VILA}: Learning image aesthetics from user comments with vision-language pretraining,'' in \emph{Proceedings of the IEEE/CVF Conference on Computer Vision and Pattern Recognition}, 2023, pp. 10\,041--10\,051.

\bibitem{yang2021reference}
N.~Yang, Q.~Zhong, K.~Li, R.~Cong, Y.~Zhao, and S.~Kwong, ``A reference-free underwater image quality assessment metric in frequency domain,'' \emph{Signal Processing: Image Communication}, vol.~94, p. 116218, 2021.

\bibitem{jiang2022underwater}
Q.~Jiang, Y.~Gu, C.~Li, R.~Cong, and F.~Shao, ``Underwater image enhancement quality evaluation: Benchmark dataset and objective metric,'' \emph{IEEE Transactions on Circuits and Systems for Video Technology}, vol.~32, no.~9, pp. 5959--5974, 2022.

\bibitem{zhou2022learning}
K.~Zhou, J.~Yang, C.~C. Loy, and Z.~Liu, ``Learning to prompt for vision-language models,'' \emph{International Journal of Computer Vision}, vol. 130, no.~9, pp. 2337--2348, 2022.

\bibitem{zhou2022conditional}
K.~\vspace{0mm}Zhou, J.~Yang, C.~C. Loy, and Z.~Liu, ``Conditional prompt learning for vision-language models,'' in \emph{Proceedings of the IEEE/CVF conference on computer vision and pattern recognition}, 2022, pp. 16\,816--16\,825.

\bibitem{chen2020simple}
T.~Chen, S.~Kornblith, M.~Norouzi, and G.~Hinton, ``A simple framework for contrastive learning of visual representations,'' in \emph{International conference on machine learning}.\hskip 1em plus 0.5em minus 0.4em\relax PMLR, 2020, pp. 1597--1607.

\bibitem{grill2020bootstrap}
J.-B. Grill, F.~Strub, F.~Altch{\'e}, C.~Tallec, P.~Richemond, E.~Buchatskaya, C.~Doersch, B.~Avila~Pires, Z.~Guo, M.~Gheshlaghi~Azar \emph{et~al.}, ``Bootstrap your own latent-a new approach to self-supervised learning,'' \emph{Advances in neural information processing systems}, vol.~33, pp. 21\,271--21\,284, 2020.

\bibitem{he2020momentum}
K.~He, H.~Fan, Y.~Wu, S.~Xie, and R.~Girshick, ``Momentum contrast for unsupervised visual representation learning,'' in \emph{Proceedings of the IEEE/CVF conference on computer vision and pattern recognition}, 2020, pp. 9729--9738.

\bibitem{bengio2009curriculum}
Y.~Bengio, J.~Louradour, R.~Collobert, and J.~Weston, ``Curriculum learning,'' in \emph{Proceedings of the 26th annual international conference on machine learning}, 2009, pp. 41--48.

\bibitem{huang2023contrastive}
S.~Huang, K.~Wang, H.~Liu, J.~Chen, and Y.~Li, ``Contrastive semi-supervised learning for underwater image restoration via reliable bank,'' in \emph{Proceedings of the IEEE/CVF Conference on Computer Vision and Pattern Recognition}, 2023, pp. 18\,145--18\,155.

\bibitem{han2022underwater}
J.~Han, M.~Shoeiby, T.~Malthus, E.~Botha, J.~Anstee, S.~Anwar, R.~Wei, M.~A. Armin, H.~Li, and L.~Petersson, ``Underwater image restoration via contrastive learning and a real-world dataset,'' \emph{Remote Sensing}, vol.~14, no.~17, p. 4297, 2022.

\bibitem{rao2022denseclip}
Y.~Rao, W.~Zhao, G.~Chen, Y.~Tang, Z.~Zhu, G.~Huang, J.~Zhou, and J.~Lu, ``Denseclip: Language-guided dense prediction with context-aware prompting,'' in \emph{Proceedings of the IEEE/CVF Conference on Computer Vision and Pattern Recognition}, 2022, pp. 18\,082--18\,091.

\bibitem{khurana2023natural}
D.~Khurana, A.~Koli, K.~Khatter, and S.~Singh, ``Natural language processing: State of the art, current trends and challenges,'' \emph{Multimedia tools and applications}, vol.~82, no.~3, pp. 3713--3744, 2023.

\bibitem{li2022human}
M.~Li, Y.~Lin, L.~Shen, Z.~Wang, K.~Wang, and Z.~Wang, ``Human perceptual quality driven underwater image enhancement framework,'' \emph{IEEE Transactions on Geoscience and Remote Sensing}, vol.~60, pp. 1--15, 2022.

\bibitem{li1906fusion}
H.~Li, J.~Li, and W.~Wang, ``A fusion adversarial underwater image enhancement network with a public test dataset. arxiv 2019,'' \emph{arXiv preprint arXiv:1906.06819}.

\bibitem{zhang2018unreasonable}
R.~Zhang, P.~Isola, A.~A. Efros, E.~Shechtman, and O.~Wang, ``The unreasonable effectiveness of deep features as a perceptual metric,'' in \emph{Proceedings of the IEEE conference on computer vision and pattern recognition}, 2018, pp. 586--595.

\bibitem{yang2015underwater}
M.~Yang and A.~Sowmya, ``An underwater color image quality evaluation metric,'' \emph{IEEE Transactions on Image Processing}, vol.~24, no.~12, pp. 6062--6071, 2015.

\bibitem{panetta2015human}
K.~Panetta, C.~Gao, and S.~Agaian, ``Human-visual-system-inspired underwater image quality measures,'' \emph{IEEE Journal of Oceanic Engineering}, vol.~41, no.~3, pp. 541--551, 2015.

\bibitem{drews2013transmission}
P.~Drews, E.~Nascimento, F.~Moraes, S.~Botelho, and M.~Campos, ``Transmission estimation in underwater single images,'' in \emph{Proceedings of the IEEE international conference on computer vision workshops}, 2013, pp. 825--830.

\bibitem{zhang2022underwater}
W.~Zhang, P.~Zhuang, H.-H. Sun, G.~Li, S.~Kwong, and C.~Li, ``Underwater image enhancement via minimal color loss and locally adaptive contrast enhancement,'' \emph{IEEE Transactions on Image Processing}, vol.~31, pp. 3997--4010, 2022.

\bibitem{islam2020fast}
M.~J. Islam, Y.~Xia, and J.~Sattar, ``Fast underwater image enhancement for improved visual perception,'' \emph{IEEE Robotics and Automation Letters}, vol.~5, no.~2, pp. 3227--3234, 2020.

\bibitem{wu2020subjective}
Q.~Wu, L.~Wang, K.~N. Ngan, H.~Li, F.~Meng, and L.~Xu, ``Subjective and objective de-raining quality assessment towards authentic rain image,'' \emph{IEEE Transactions on Circuits and Systems for Video Technology}, vol.~30, no.~11, pp. 3883--3897, 2020.

\bibitem{10155564}
R.~Cong, W.~Yang, W.~Zhang, C.~Li, C.-L. Guo, Q.~Huang, and S.~Kwong, ``{PUGAN}: Physical model-guided underwater image enhancement using gan with dual-discriminators,'' \emph{IEEE Transactions on Image Processing}, vol.~32, pp. 4472--4485, 2023.

\bibitem{2020SVAM}
M.~J. Islam, R.~Wang, and J.~Sattar, ``{SVAM}: Saliency-guided visual attention modeling by autonomous underwater robots,'' \emph{arXiv preprint arXiv:2011.06252}, 2020.

\bibitem{9881581}
T.~Ren, H.~Xu, G.~Jiang, M.~Yu, X.~Zhang, B.~Wang, and T.~Luo, ``Reinforced swin-convs transformer for simultaneous underwater sensing scene image enhancement and super-resolution,'' \emph{IEEE Transactions on Geoscience and Remote Sensing}, vol.~60, pp. 1--16, 2022.

\bibitem{10251985}
H.~Yan, Z.~Zhang, J.~Xu, T.~Wang, P.~An, A.~Wang, and Y.~Duan, ``{UW-CycleGAN}: Model-driven cyclegan for underwater image restoration,'' \emph{IEEE Transactions on Geoscience and Remote Sensing}, vol.~61, pp. 1--17, 2023.

\bibitem{10378578}
Z.~Liang, C.~Li, S.~Zhou, R.~Feng, and C.~C. Loy, ``Iterative prompt learning for unsupervised backlit image enhancement,'' in \emph{2023 IEEE/CVF International Conference on Computer Vision (ICCV)}, 2023, pp. 8060--8069.

\bibitem{fu2022unsupervised}
Z.~Fu, H.~Lin, Y.~Yang, S.~Chai, L.~Sun, Y.~Huang, and X.~Ding, ``Unsupervised underwater image restoration: From a homology perspective,'' in \emph{Proceedings of the AAAI Conference on Artificial Intelligence}, vol.~36, no.~1, 2022, pp. 643--651.

\bibitem{hummel1975image}
R.~Hummel, ``Image enhancement by histogram transformation,'' \emph{Unknown}, 1975.

\bibitem{9749233}
Q.~Jiang, Y.~Gu, C.~Li, R.~Cong, and F.~Shao, ``Underwater image enhancement quality evaluation: Benchmark dataset and objective metric,'' \emph{IEEE Transactions on Circuits and Systems for Video Technology}, vol.~32, no.~9, pp. 5959--5974, 2022.

\bibitem{lin2024mirrordiffusion}
Y.~Lin, X.~Xian, Y.~Shi, and L.~Lin, ``{MirrorDiffusion}: Stabilizing diffusion process in zero-shot image translation by prompts redescription and beyond,'' \emph{IEEE Signal Processing Letters}, 2024.

\bibitem{lin2022exploring}
Y.~Lin, S.~Zhang, T.~Chen, Y.~Lu, G.~Li, and Y.~Shi, ``Exploring negatives in contrastive learning for unpaired image-to-image translation,'' in \emph{Proceedings of the 30th ACM International Conference on Multimedia}, 2022, pp. 1186--1194.

\bibitem{johnson2016perceptual}
J.~Johnson, A.~Alahi, and L.~Fei-Fei, ``Perceptual losses for real-time style transfer and super-resolution,'' in \emph{Computer Vision--ECCV 2016: 14th European Conference, Amsterdam, The Netherlands, October 11-14, 2016, Proceedings, Part II 14}.\hskip 1em plus 0.5em minus 0.4em\relax Springer, 2016, pp. 694--711.

\bibitem{paull2013auv}
L.~Paull, S.~Saeedi, M.~Seto, and H.~Li, ``{AUV} navigation and localization: A review,'' \emph{IEEE Journal of oceanic engineering}, vol.~39, no.~1, pp. 131--149, 2013.

\bibitem{cong2021rrnet}
R.~Cong, Y.~Zhang, L.~Fang, J.~Li, Y.~Zhao, and S.~Kwong, ``{RRNet}: Relational reasoning network with parallel multiscale attention for salient object detection in optical remote sensing images,'' \emph{IEEE Transactions on Geoscience and Remote Sensing}, vol.~60, pp. 1--11, 2021.

\bibitem{wang2023reinforcement}
H.~Wang, S.~Sun, X.~Bai, J.~Wang, and P.~Ren, ``A reinforcement learning paradigm of configuring visual enhancement for object detection in underwater scenes,'' \emph{IEEE Journal of Oceanic Engineering}, vol.~48, no.~2, pp. 443--461, 2023.

\bibitem{li2019nested}
C.~Li, R.~Cong, J.~Hou, S.~Zhang, Y.~Qian, and S.~Kwong, ``Nested network with two-stream pyramid for salient object detection in optical remote sensing images,'' \emph{IEEE Transactions on Geoscience and Remote Sensing}, vol.~57, no.~11, pp. 9156--9166, 2019.

\bibitem{anwar2020diving}
S.~Anwar and C.~Li, ``Diving deeper into underwater image enhancement: A survey,'' \emph{Signal Processing: Image Communication}, vol.~89, p. 115978, 2020.

\bibitem{9825662}
Z.~Huang, J.~Li, Z.~Hua, and L.~Fan, ``Underwater image enhancement via adaptive group attention-based multiscale cascade transformer,'' \emph{IEEE Transactions on Instrumentation and Measurement}, vol.~71, pp. 1--18, 2022.

\bibitem{10192442}
C.~Liu, X.~Shu, L.~Pan, J.~Shi, and B.~Han, ``Multiscale underwater image enhancement in rgb and hsv color spaces,'' \emph{IEEE Transactions on Instrumentation and Measurement}, vol.~72, pp. 1--14, 2023.

\end{thebibliography}


\newpage

 






\vfill

\end{document}